\documentclass[letterpaper]{article}
\usepackage{aaai23} %
\usepackage{times} %
\usepackage{helvet} %
\usepackage{courier} %
\usepackage[hyphens]{url} %
\usepackage{graphicx} %
\usepackage{graphicx} %
\usepackage{natbib} %
\usepackage{caption} %
\usepackage{algorithm}
\usepackage{algorithmic}

\usepackage{newfloat}
\usepackage{listings}
\DeclareCaptionStyle{ruled}{labelfont=normalfont,labelsep=colon,strut=off} %
\floatstyle{ruled}
\newfloat{listing}{tb}{lst}{}
\floatname{listing}{Listing}
\usepackage[utf8]{inputenc} %
\usepackage[T1]{fontenc}    %
\usepackage{url}            %
\usepackage{amsfonts}       %
\usepackage{amsmath}
\usepackage{nicefrac}       %
\usepackage{microtype}      %
\usepackage{xcolor}         %
\usepackage{graphicx}
\usepackage{enumitem}
\usepackage{multicol}
\usepackage{tabularx}
\usepackage{etoolbox}
\newtoggle{inlinecomments}
\togglefalse{inlinecomments}  %
\iftoggle{inlinecomments}{
\newcommand{\dave}[1]{{\color{orange} [DM[#1]]}}
\newcommand{\shumeet}[1]{{\color{green} [SB[#1]]}}
\newcommand{\yairmov}[1]{{\color{blue} [YA[#1]]}}
\newcommand{\saurabh}[1]{{\color{red} [SS[#1]]}}
}{
\newcommand{\dave}[1]{}
\newcommand{\shumeet}[1]{}
\newcommand{\yairmov}[1]{}
\newcommand{\saurabh}[1]{}
}
\DeclareMathOperator{\codiff}{codiff}
\newcommand\variablename[1]{\mathop{\textnormal{\slshape #1}}\nolimits}

\newcommand{\CAMATH}{\variablename{CA}}
\title{Diversity and Diffusion:\\Observations on Synthetic Image Distributions with Stable Diffusion}

\author{
    David Marwood, Shumeet Baluja, Yair Alon\\
}
\affiliations {
    Google Research\\
    \{marwood, shumeet, yairmov\}@google.com
}

\newcommand{\tti}{TTI~}
\newcommand{\ttins}{TTI}

\newcommand{\cd}{Centroid Distance }

\newcommand{\sd}{SD~}
\newcommand{\sdns}{SD}
\newcommand{\knn}{\emph{k-}nn~}

\newcommand{\meanshift}{centroid-shift}

\newcommand{\eg} {\emph{e.g.~}}
\newcommand{\random} {~$\overset{{\scriptscriptstyle \operatorname{R}}}{\leftarrow}$~}

\begin{document}
\maketitle

\begin{abstract}

Recent progress in text-to-image (TTI) systems, such as Stable Diffusion, Imagen, and DALL-E 2, have made it possible to create realistic images with simple text prompts.  It is tempting to use these systems to eliminate the manual task of obtaining natural images for training a new machine learning classifier.  However, in all of the experiments performed to date, classifiers trained solely with synthetic images perform poorly at inference, despite the images used for training appearing realistic.  Examining this apparent incongruity in detail gives insight into the limitations of the underlying image generation processes.  Through the lens of diversity in image creation vs. accuracy of what is created,  we dissect the differences in semantic mismatches in what is modeled in synthetic vs. natural images. This will elucidate the roles of the image-language model, CLIP, and the image generation model, diffusion. We find four issues that limit the usefulness of TTI systems for this task: ambiguity, adherence to prompt, lack of diversity, and inability to represent the underlying concept. We further present surprising insights into the geometry of CLIP embeddings.

\end{abstract}

\section{Introduction}
Large text-to-image (TTI) models have visually demonstrated the remarkable recent progress in AI, enabling high-quality synthesis of images from textual prompts.  The most commonly used successful models employ diffusion, a gradual denoising process \cite{Ho2020DDPM}.  These diffusion-based generative text-to-image models \cite{imagen2022, rombach2021highresolution, chang2023muse, yu2022scaling, ramesh2022hierarchical, nichol2022glide} have seen impressive improvements in quality that rival or exceed modern Generative Adversarial Networks (GAN) techniques \cite{kang2023gigagan, dhariwal2021beatgans}. 

\begin{figure}
\centering

\setlength{\tabcolsep}{1pt}
\setlength\extrarowheight{-10pt}
\begin{tabular}{cccccc}

\includegraphics[width=0.072\textwidth]{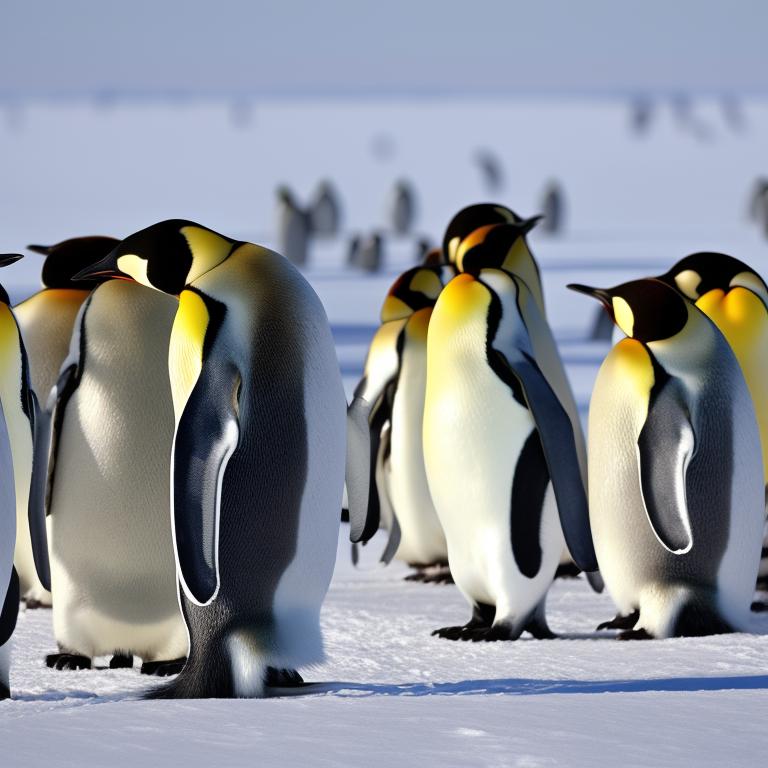}&
\includegraphics[width=0.072\textwidth]{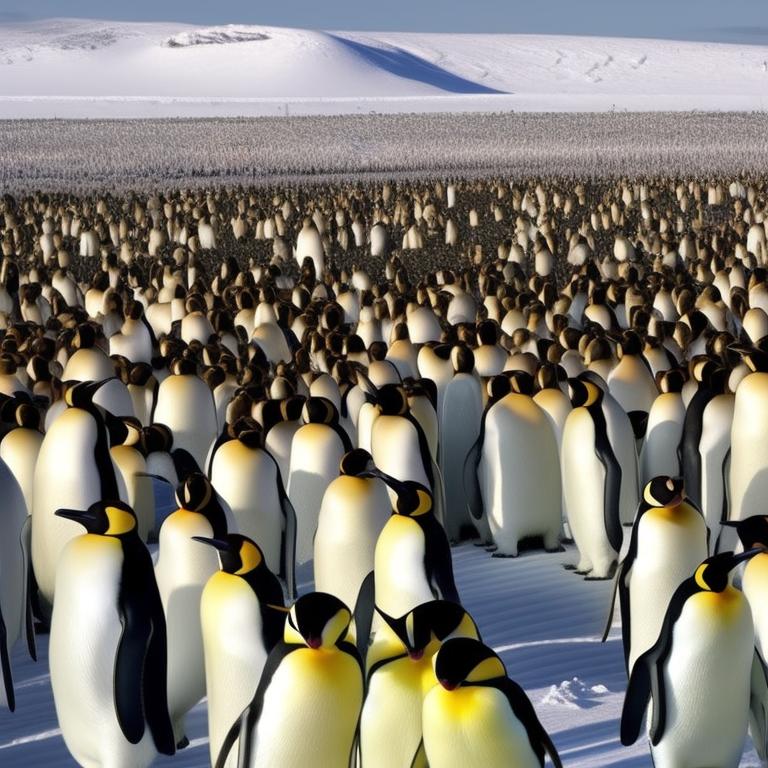}&
\includegraphics[width=0.072\textwidth]{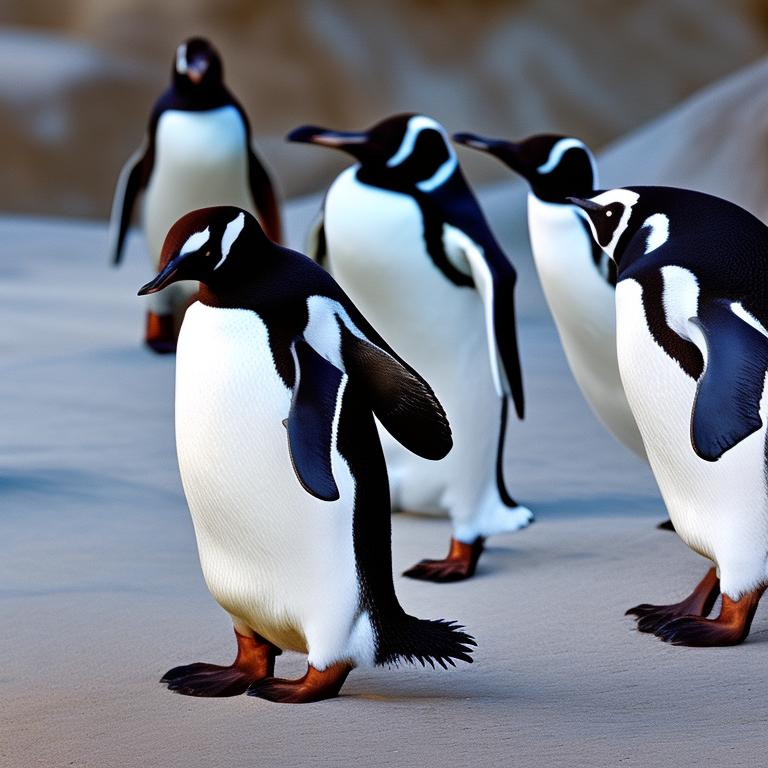}&
\includegraphics[width=0.072\textwidth]{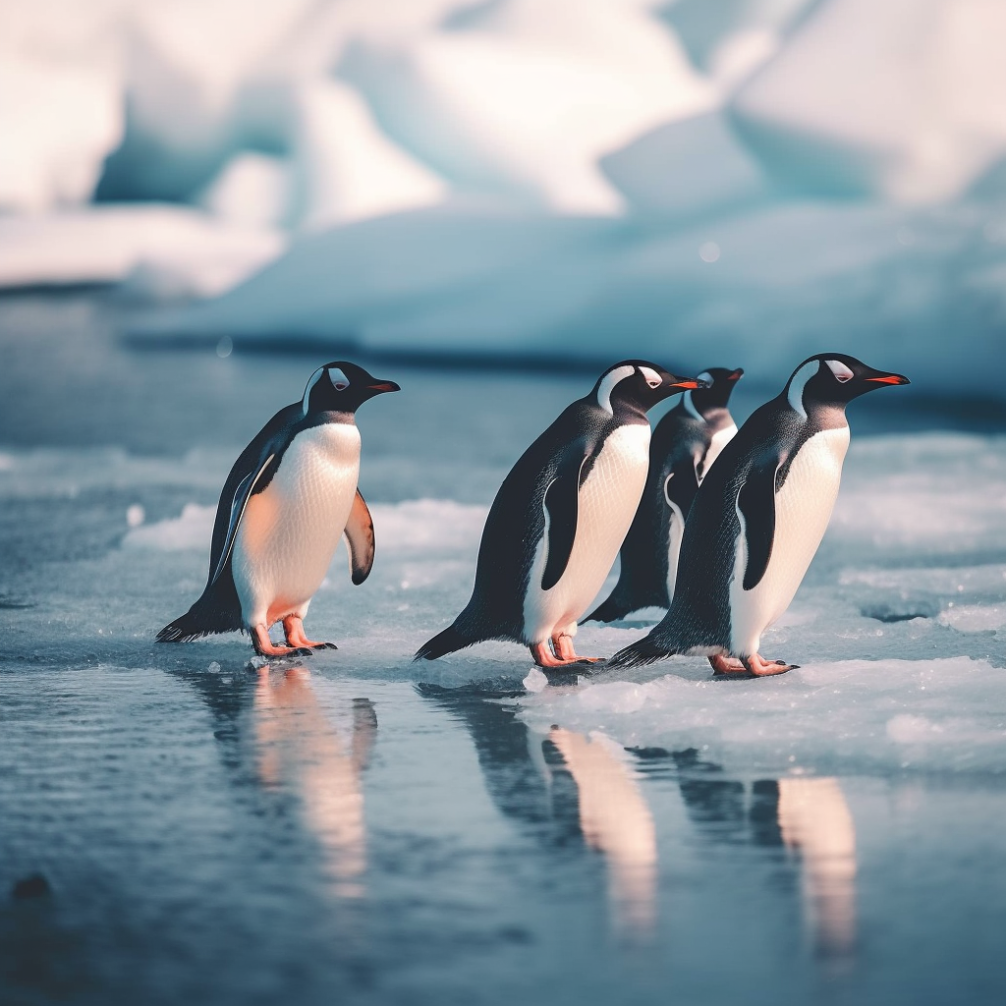}&
\includegraphics[width=0.072\textwidth]{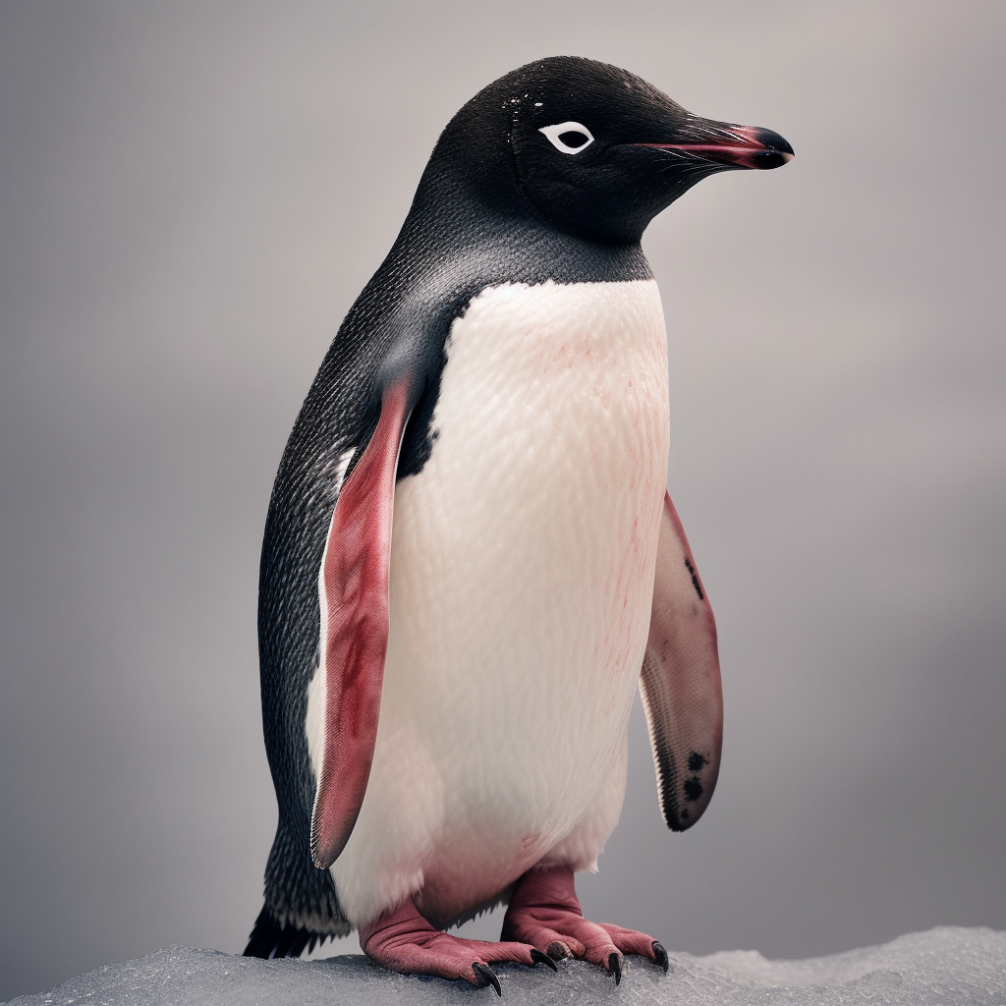}&
\includegraphics[width=0.072\textwidth]{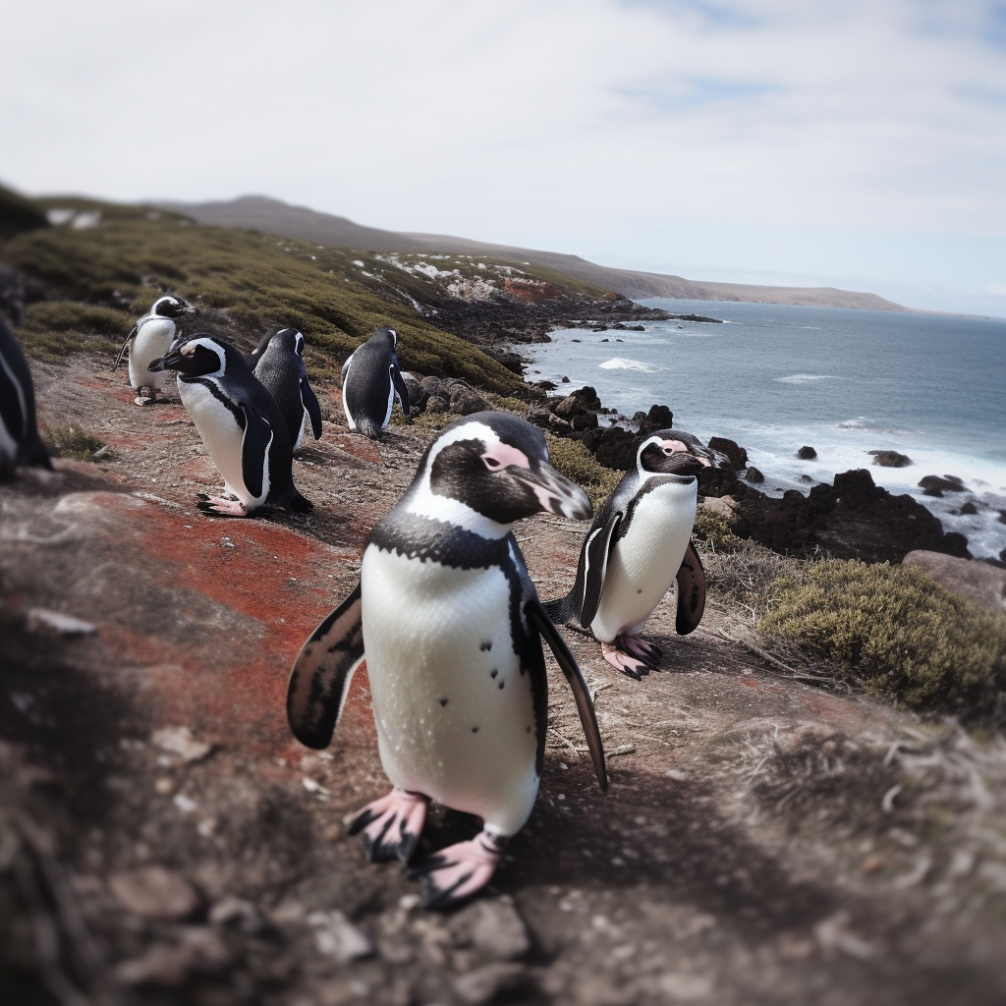}\\
\includegraphics[width=0.072\textwidth]{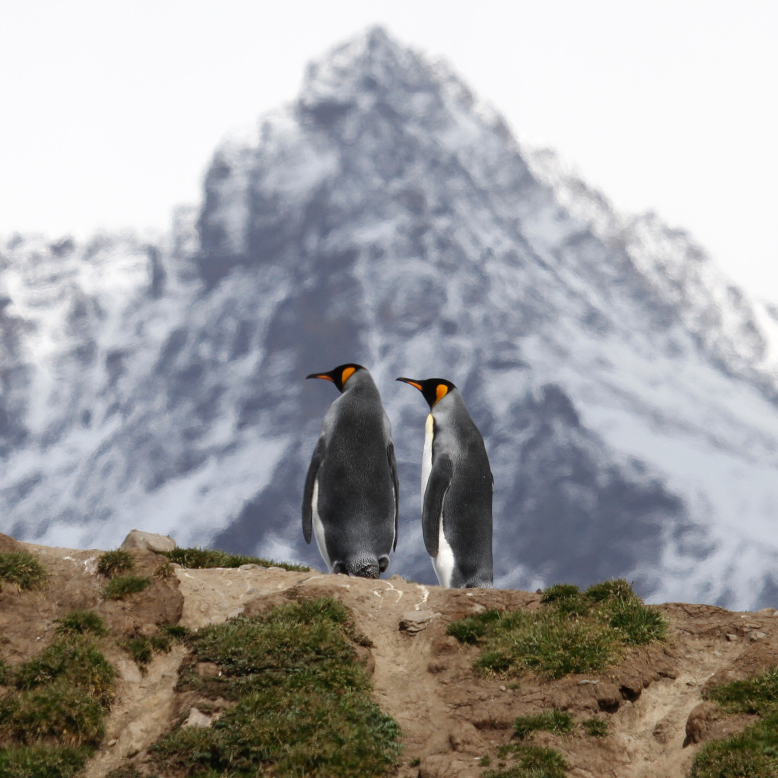}&
\includegraphics[width=0.072\textwidth]{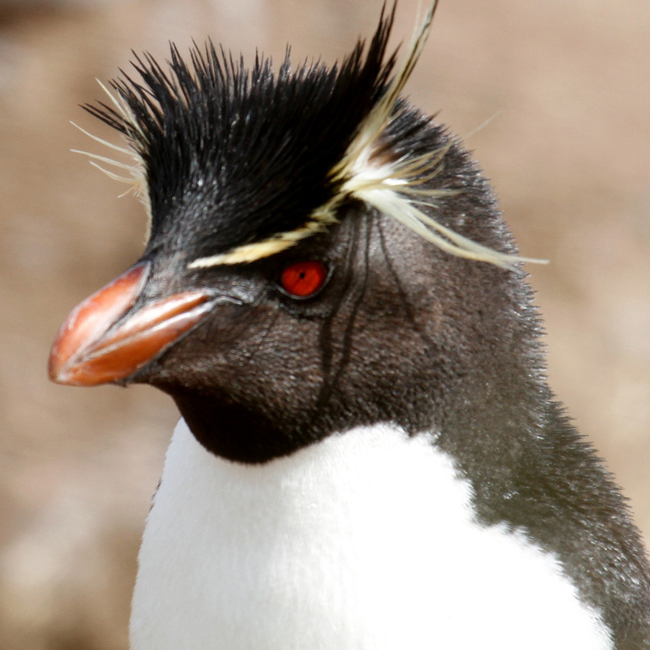}&
\includegraphics[width=0.072\textwidth]{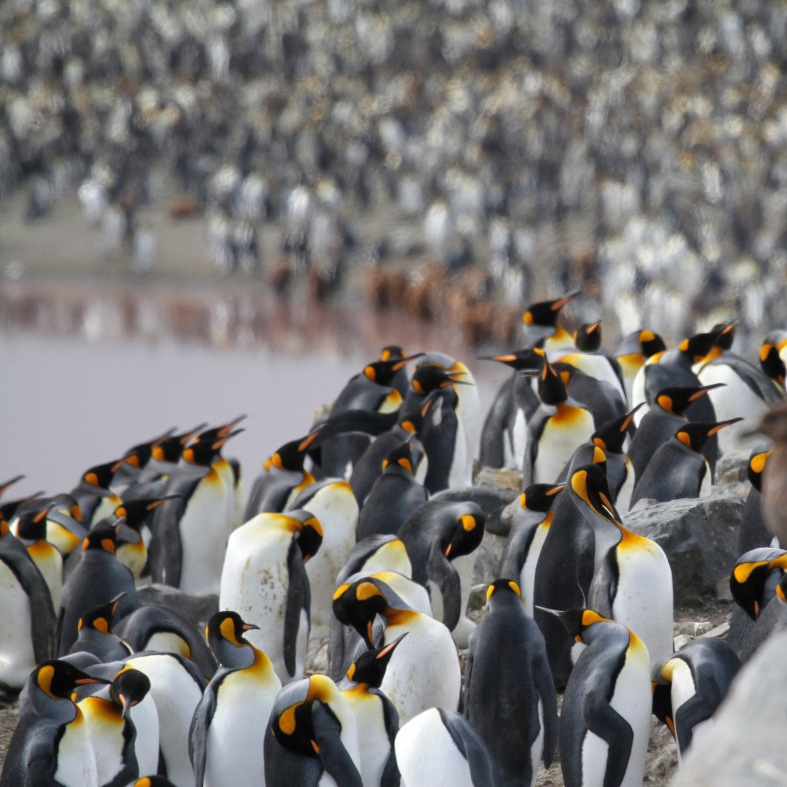}&
\includegraphics[width=0.072\textwidth]{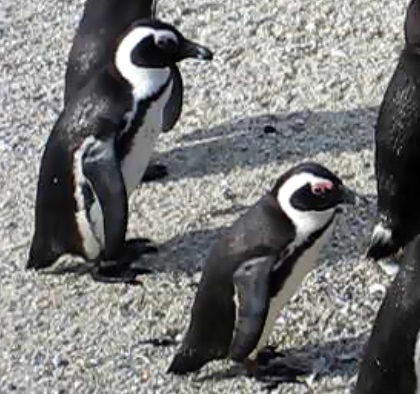}&
\includegraphics[width=0.072\textwidth]{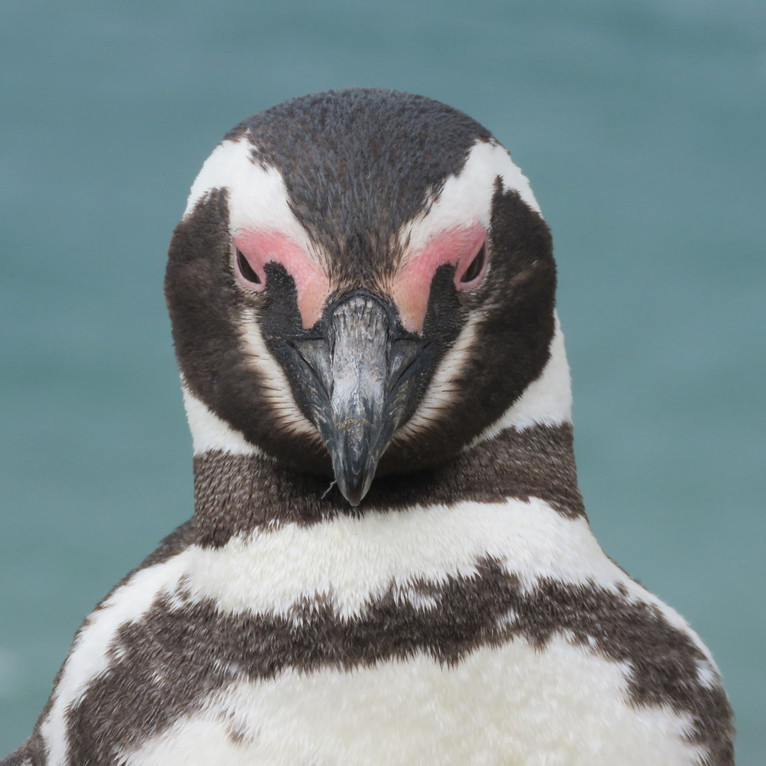}&
\includegraphics[width=0.072\textwidth]{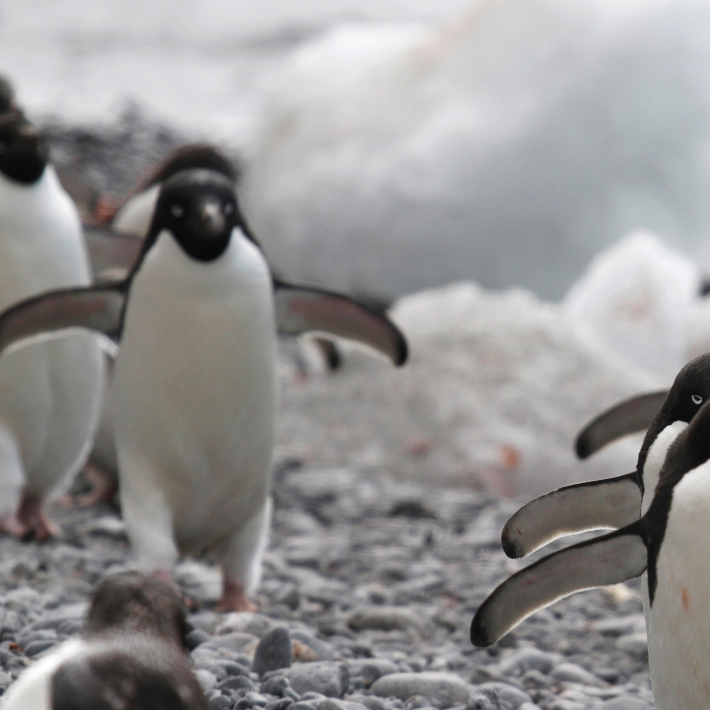}\\
\end{tabular}

   \caption{Penguins.  Top Row: Synthetic Images.  Bottom Row: Natural photographs from ~\cite{inaturalist}. 
   }
  \label{fig:teaser}    

\end{figure}

As generative models continue to produce increasingly realistic images, beyond creating images for artistic or more general human viewing, we can ask whether synthetic images can replace traditional datasets gathered by hand.  To an untrained eye, the images often appear to be suitable; see Figure~\ref{fig:teaser}.  If so, creating datasets, even ones much larger than today's datasets, can become a far less manual process. Conceivably, it is possible that the generative models could produce even more diverse images than the distribution of a hand-gathered set, potentially exceeding their utility.

Multiple works have explored exactly this avenue and have trained classifiers with synthetic-only image sets, typically using ImageNet-1K \cite{Deng2009ImageNet} classification as a benchmark \cite{azizi2023synthetic, sariyildiz2023fake, bansal2023leaving}.  Interestingly, none of these studies, using purely synthetic images, matches, or even comes close, to the accuracy of training with natural images.  Instead, they combine synthetic and natural images to improve accuracy. Nonetheless, even then,  a too large proportion of synthetic images degrades the performance; the number of natural images required increases with the number of synthetic images, only increasing the manual process.

Our primary contribution is to uncover the root causes of the poor performance with purely synthetic images.  Given the nascent nature of this field, even choosing which aspects to study is open for discussion. Here, we examine the roles of diversity and meaning-shift in both the internal representations and the generation procedures used inside a \tti system.  
To make our work as  widely applicable as possible, we study Stable Diffusion's synthesized images in comparison to ImageNet's natural images.

We present related work in Section~\ref{sec:related-work}. Section~\ref{sec:reasons} gives four reasons for the drop in accuracy using synthetic images.
Section~\ref{sec:modified-prompts} examines the geometry of prompt and image embeddings created by CLIP.  Finally, recognizing that zero-shot classifiers are equivalent to \knn classifiers with class centroids, we look at  \knn classifiers broadly.

\section{Closely Related Work}
\label{sec:related-work}
\textbf{Synthetic Images for Classification.} Foundational to our study are recent attempts at ImageNet classification training using synthetic images. Using the short names for ImageNet-1K classes, or prompts derived from them, a \tti system, such as Stable Diffusion, is used to generate thousands of images for that class that are then used for training. While the ImageNet set has approximately 1,000 images per class, synthetic sets are not limited to this. All of the previous studies used techniques to improve the synthetic image prompts that led to accuracy improvements over pure short-names.  Though some studies could exceed their natural-image-only benchmark on subsets of ImageNet, we constrain our study to Top-1 accuracy on the full ImageNet validation set.
For the full test, all of the studies see a very large gap (often more than 30\%) between training on a natural images  and synthetic ones~\cite{sariyildiz2023fake, azizi2023synthetic, Xie2016ResNext}.  As a middle ground, they all explore forms of fine-tuning or data augmentation for natural images to improve baseline accuracies.  However, this technique still demonstrates a continued reliance on hand-gathered data.

\textbf{Diffusion Models.} Many generative \tti systems use diffusion~\cite{Ho2020DDPM, Song2021DDIM}, including Stable Diffusion (SD), Imagen, DALL-E 2 \cite{ramesh2022hierarchical}, Openjourney \cite{Openjourney}, Versatile Diffusion \cite{xu2023versatile} and Glide \cite{nichol2022glide}. Diffusion has been used in other generative domains such as audio synthesis \cite{Kong2021DiffWave} and video \cite{chen2023controlavideo}. Our analysis investigates the contribution of diffusion to accuracy; however, the ideas trivially generalize to non-diffusion \tti systems like Cogview \cite{ding2021cogview}, Parti \cite{yu2022scaling} and Muse \cite{chang2023muse}.   Additionally, we note that methods of fine-tuning diffusion, such as Dreambooth and textual inversion \cite{ruiz2023dreambooth, gal2022image, tewel2023keylocked} can be used to better create images similar to ImageNet classes.   This yields improvements,  but again requires the use of an existing image database.

\textbf{Synthetic Data from GANs.} Like diffusion-based-\ttins, Generative Adversarial Networks (GANs) \cite{goodfellow2020generative} can create synthetic training examples without hand-implemented domain knowledge, for example for facial expression classification \cite{Bhattarai2020SampleStrategies}, traffic sign classification \cite{Dewi2021TrafficSign} and semantic segmentation \cite{Sankaranarayanan2018GANSegmentation}. Such synthetic data for training are typically labeled in the tasks domain while the diffusion models are largely classifier-free.   For the wider task of broad text-based synthesis, \tti systems have often shown more fidelity to natural images \cite{yu2022scaling}.

\section{The Pitfalls of Purely Synthetic Data}
\label{sec:reasons}
Synthetic images are poor replacements for natural images when training classifiers.  We offer four explanations for this.

\subsection{Reason \#1: Prompt and Class Ambiguity}
\label{sec:ambiguous}
Many short names \cite{ImageNetLabelsAnisha} for the ImageNet-1K classes can be inadequate  for use with a \tti system.  First, the ImageNet short names are often less precise than their associated WordNet \cite{Miller1995WordNet}
synset description.
For example, the synset ``African chameleon, Chamaeleo chamaeleon'' is more precise than the ImageNet short name, ``chameleon.''   In contrast to some previous studies, we  avoid these issues by starting with synset descriptions as prompts. 

\begin{figure}
\centering
\small
\setlength{\tabcolsep}{1pt}
\begin{tabular}{cccccc}

\multicolumn{2}{c}{``drake''} & \multicolumn{1}{c|}{~} & & \multicolumn{2}{c}{``whiskey jug''} \\
ImageNet & \multicolumn{1}{|c}{SD} & \multicolumn{1}{c|}{~} & & \multicolumn{1}{c|}{ImageNet} & SD \\
\hline
\hline
\includegraphics[width=1.5cm]{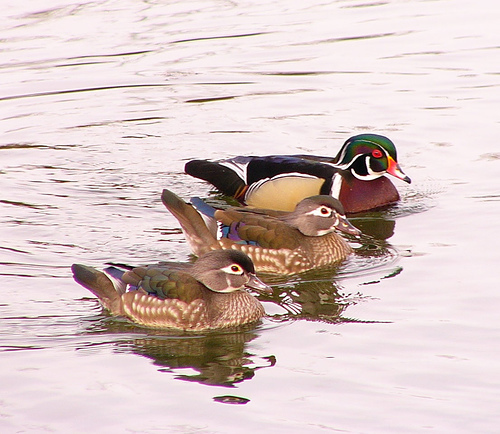} &
\includegraphics[height=1.5cm]{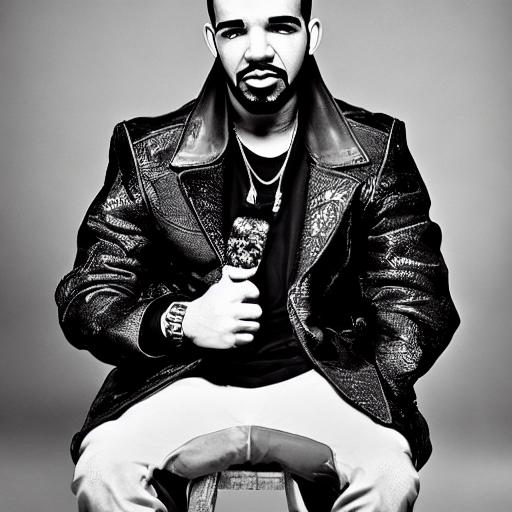} &
~&~& 
\includegraphics[height=1.5cm]{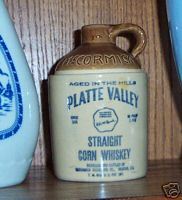} &
\includegraphics[height=1.5cm]{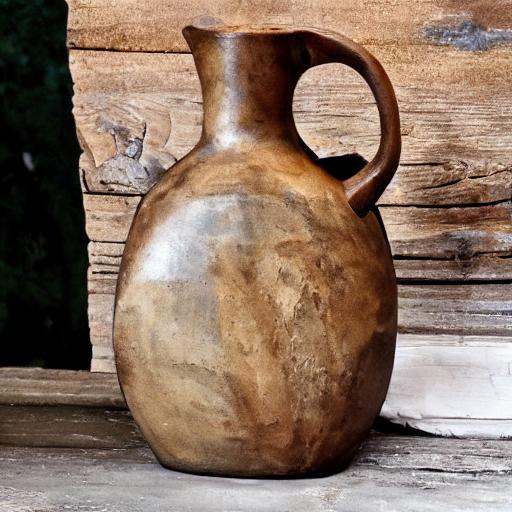} \\
\includegraphics[width=1.5cm]{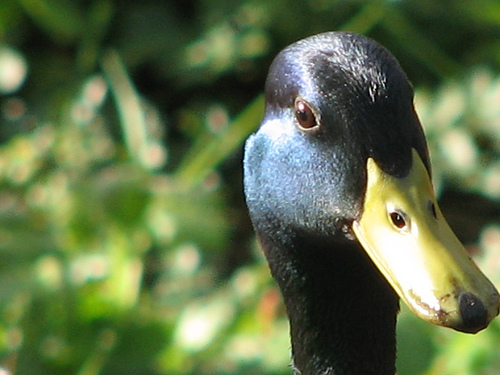} &
\includegraphics[height=1.5cm]{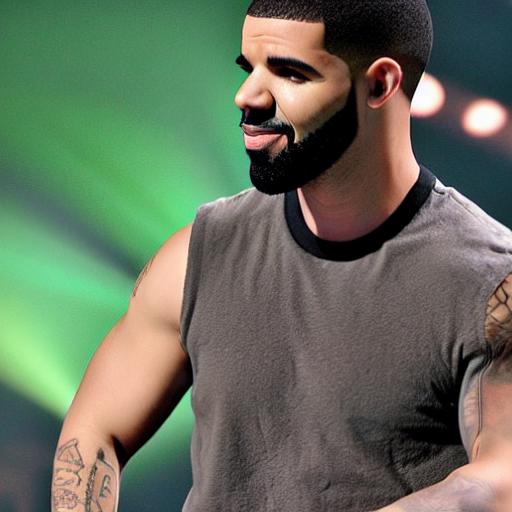} &
~&~& 
\includegraphics[height=1.5cm]{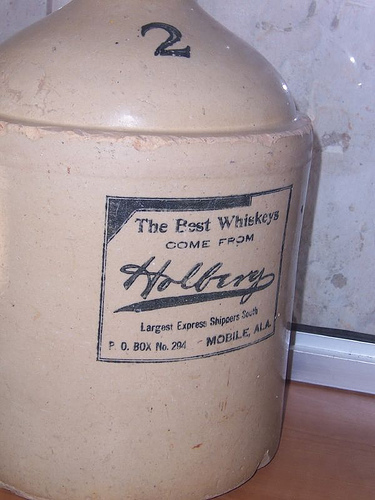} &
\includegraphics[height=1.5cm]{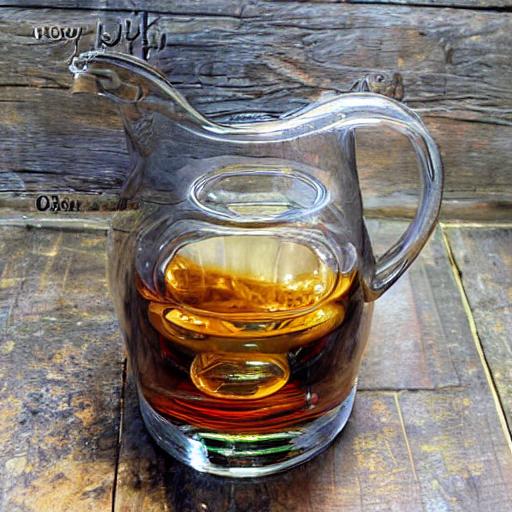} \\
\end{tabular}
   \caption{Images from  ImageNet  and generated by SD. ``drake'' confuses the duck with the homonymous performer. ``whiskey jug'' shows the bias in  ImageNet.}
  \label{fig:ambiguity}    

\end{figure}

Second, often the words in the synset may have homonyms and are therefore ambiguous.  SD may have a different meanings than the synset descriptions; it is biased to produce likely images, where ``likely'' is  influenced by popular culture. 
Figure~\ref{fig:ambiguity} shows the query ``drake'' for which ImageNet contains ducks but SD produces the homonymous performer.

Third, synsets are often repetitive, \eg the ``chameleon'' class (above), %
causing SD to produce multiple instances of the object rather than  interpreting these as synonyms.

Fourth, when humans gather datasets,  biases may be introduced in data collection.  When ImageNet is used for both training and evaluation without visual examination, these   biases remain hidden because train/evaluation  are consistent with each other.  However, when compared with the common usage of the label, the biases become evident.     For example, ImageNet's ``whiskey jug'' images are clay jars with distinctive labels, a subset of the general class of "whiskey jugs." See Figure~\ref{fig:ambiguity}.  \tti systems will not be limited to a specific type of whiskey jar and will produce out-of-domain images relative to ImageNet.  See \cite{beyer2020imagenet, Stock_2018_ECCV, pmlr-v119-tsipras20a} for a discussion of ImageNet's limitations. Unfortunately, this has often not been addressed in recent work,  and may substantially impact the results.

We broadly categorize these problems as \textbf{ambiguity}. To alleviate these issues, we generated 50 images for each label and
classified them with a ResNet-RS-152 classifier\footnote{The full training procedure follows \cite{modelGarden}.  Using the ImageNet-training set, top-1 score is 80.8\%  on ImageNet-validation.}. We manually reviewed the 300 labels with the lowest accuracy.  For classes with low accuracy that we determined to be \textit{caused solely by ambiguity}, the labels were minimally clarified.
This resulted in 105 modifications.\footnote{Provided at \url{http://anonymous.location.com}.}

Before continuing to the deeper analysis of synthetic images, we quantify the improvement of the modified prompts.   In our baseline synthetic image set, for each of the 1000 classes we created 1200 training images and 50 evaluation images (total 1,250,000 images created) using Stable Diffusion v1.5 (SD) with the original synsets as prompts. We term this set \textit{Synset}.   Next, we replaced the 105 classes 
with images created using the newly clarified prompts (131,250 images replaced); this disambiguated set is \textit{Disamb}.

We use these two sets, each with training and evaluation splits, as drop-in replacements for the original ImageNet set; see Table~\ref{table:resnet}.  
In our first test, when the standard ImageNet trained network is \emph{tested} on the Synset and Disamb sets, the performance drops to 68.5\% and 72.2\% respectively (lines 2-3). This is the first indication that (1) there is a mismatch in the images between ImageNet and the synthetic images, and (2) the modification to the 105 classes was beneficial.

\begin{table}
  \small
  \begin{center}
  \begin{tabular}{c|c|c|c}

    & Training Set & Evaluation Set & Accuracy(\%)\\ %
    \hline
    \hline
    1. & Imagenet  & Imagenet  & 80.8\\
    2. &Imagenet  & Synset &  68.5\\
    3. &Imagenet  & Disamb  &  72.2\\
    \hline
    4. &Synset  & Imagenet  &  23.0\\
    5. &Synset  & Synset &    98.5 \\
    \hline
    6. &Disamb  & Imagenet & 26.3 \\
    7. &Disamb  & Disamb & 98.8 \\

  \end{tabular}
  \caption{Accuracy of various train and evaluation sets, measured by training a ResNet-RS-152. }
  \label{table:resnet}  
  \end{center}

\end{table}

In our second test, we  use the Synset and Disamb sets for \emph{training}.  Training with Synset and evaluating on ImageNet precipitously reduces performance to 23\% (random is .1\%) (line 4).  Training with Disamb improves the results modestly to 26.3\%.   What happened?
Given that training with natural images and evaluating with synthesized images did not suffer as much degradation as training with synthesized and evaluating with natural images, perhaps the synthesized images have less diversity. The useful synthetic images may represent less breadth.  Table~\ref{table:resnet} lines 5 \& 7 support this line of inquiry:  if we both train and evaluate with synthetic images only, and they  have low diversity, the accuracies should be high.  They are: 98-99\%. This leads to Reasons \#3 \& \#4,  which will contrast \emph{diversity} with \emph{\meanshift}, in which a different concept may be represented in the images than expected.  Next, reason \#2 continues with the role of the prompt and   controlling its effect on image generation.

\subsection{Reason \#2: Adherence to the Text Prompt}
\label{sec:low-diversity}

In most \tti systems, there is an explicit parameter that controls how much the generated image should be conditioned on the given textual prompt.  In \sdns, the parameter \emph{clip\_guidance\_scale (cgs)} controls the influence of the prompt on the otherwise unconditional image generation. 
The default in SD is $cgs=7.5$, which was used for generating both Synset and Disamb sets. We hypothesized that this factor, which represents the adherence to the prompt, may be too high to allow enough diversity in synthetic images.

\begin{figure}[t]
    \centering
\small
\setlength{\tabcolsep}{1pt}

\begin{tabular}{c|cccc}
 & 1.0  & 2.5          & 5.0 & 7.5      \\
 &      & (Disamb2.5)  &     & (Disamb) \\
 \hline
 \hline

    \rotatebox[origin=c]{90}{\begin{minipage}{1.5cm}\centering
    grand piano,\\piano\end{minipage}} & 
    \raisebox{-0.45\height}{\includegraphics[height=1.5cm]{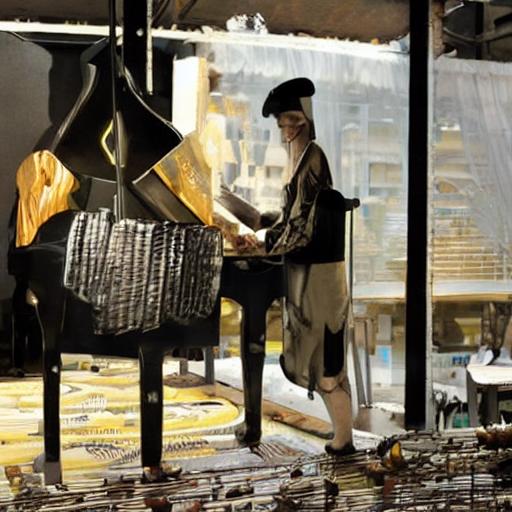}} &
    \raisebox{-0.45\height}{\includegraphics[height=1.5cm]{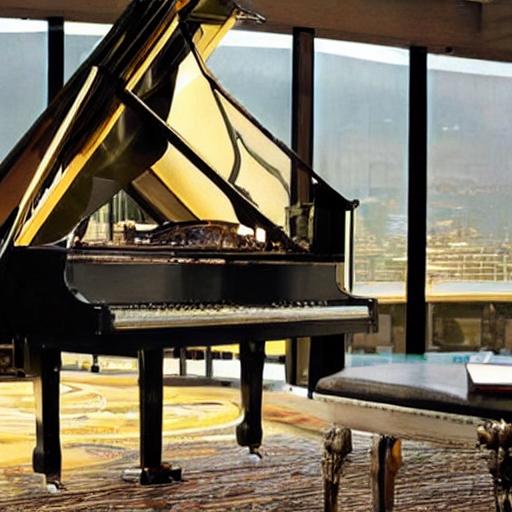}} &
    \raisebox{-0.45\height}{\includegraphics[height=1.5cm]{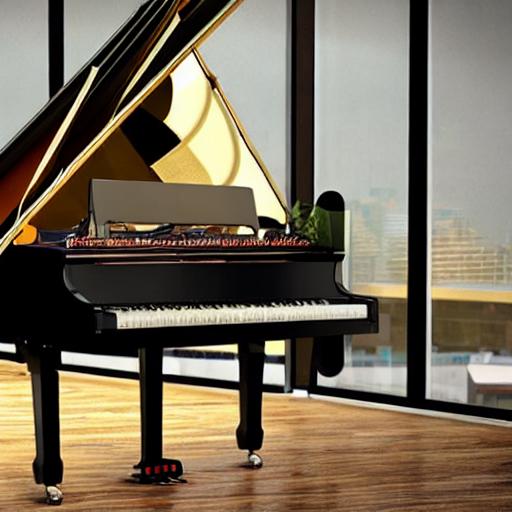}} &
    \raisebox{-0.45\height}{\includegraphics[height=1.5cm]{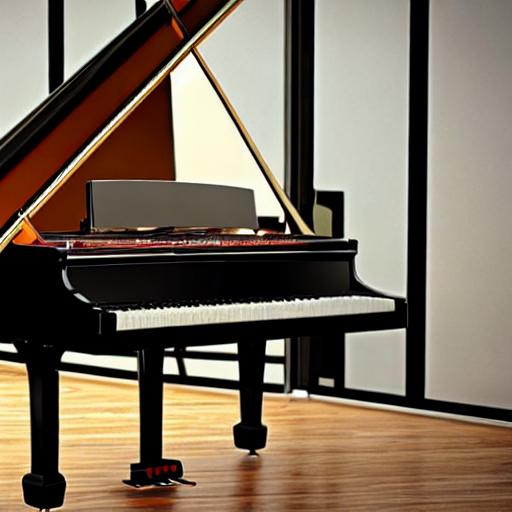}} \\

    \rotatebox[origin=c]{90}{\begin{minipage}{1.5cm}\centering
    pickup,\\pickup truck\end{minipage}} & 
    \raisebox{-0.45\height}{\includegraphics[height=1.5cm]{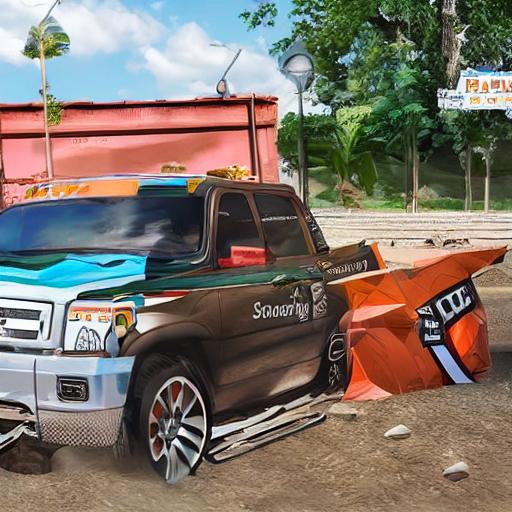}} &
    \raisebox{-0.45\height}{\includegraphics[height=1.5cm]{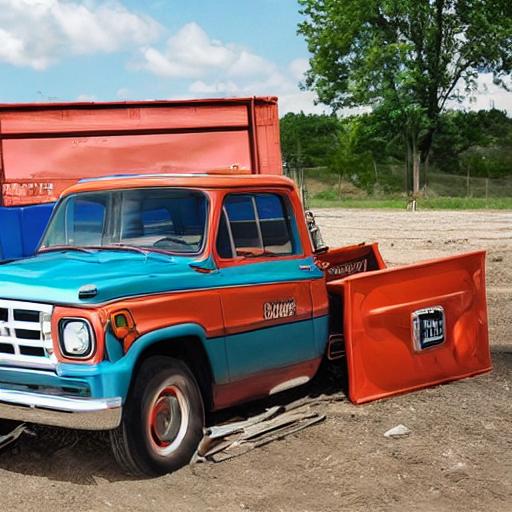}} &
    \raisebox{-0.45\height}{\includegraphics[height=1.5cm]{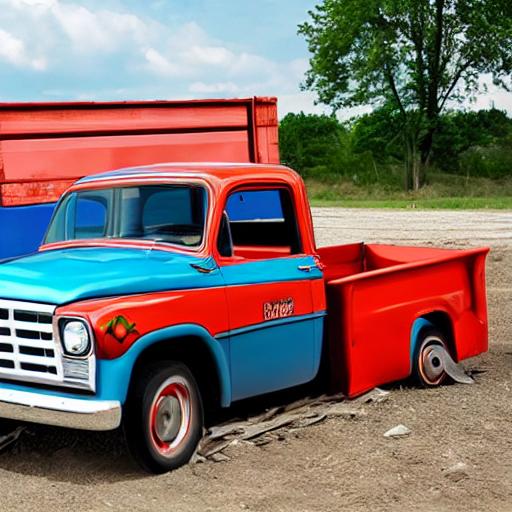}} &
    \raisebox{-0.45\height}{\includegraphics[height=1.5cm]{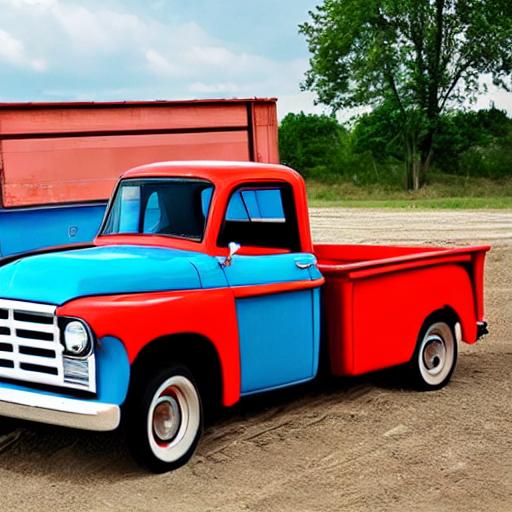}} \\

    \rotatebox[origin=c]{90}{\begin{minipage}{1.5cm}\centering
    burrito\end{minipage}} & 
    \raisebox{-0.45\height}{\includegraphics[height=1.5cm]{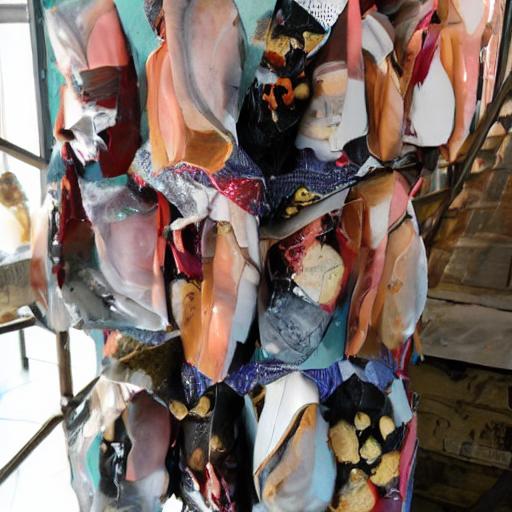}} &
    \raisebox{-0.45\height}{\includegraphics[height=1.5cm]{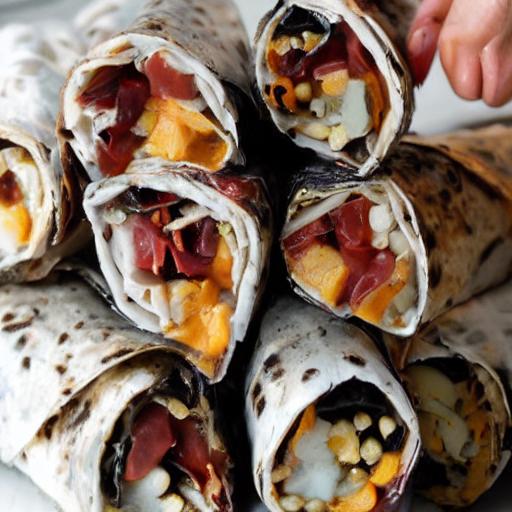}} &
    \raisebox{-0.45\height}{\includegraphics[height=1.5cm]{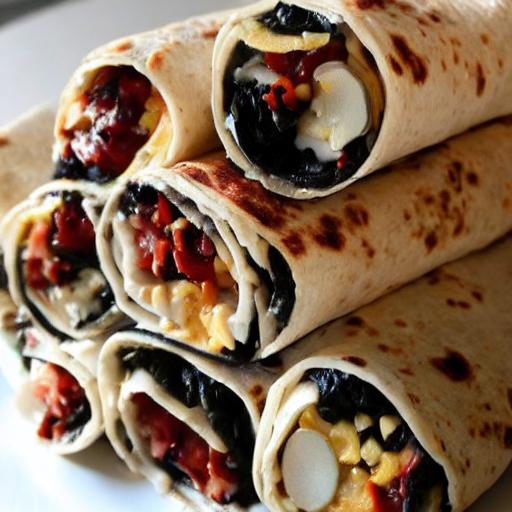}} &
    \raisebox{-0.45\height}{\includegraphics[height=1.5cm]{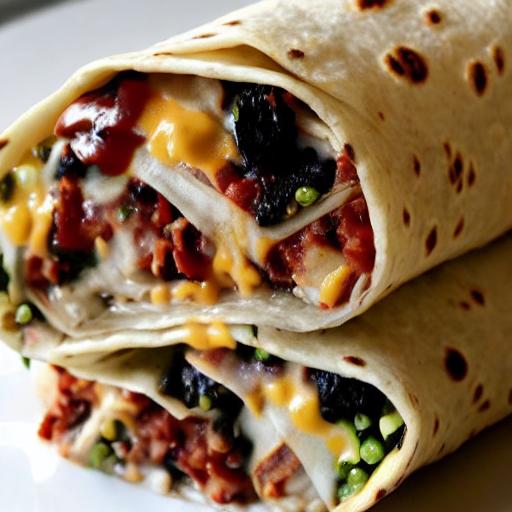}} \\

    \rotatebox[origin=c]{90}{\begin{minipage}{1.5cm}\centering
    green\\mamba\end{minipage}} & 
    \raisebox{-0.45\height}{\includegraphics[height=1.5cm]{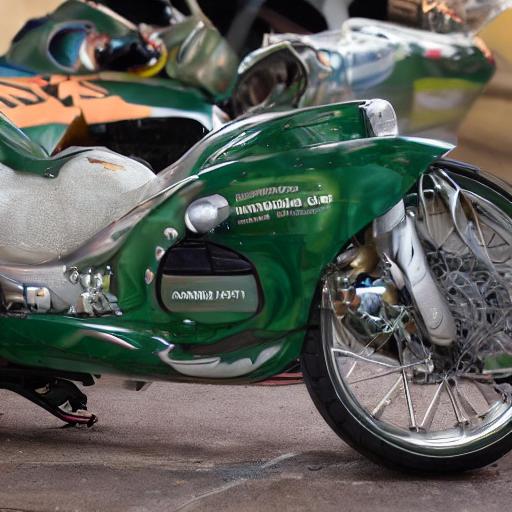}} &
    \raisebox{-0.45\height}{\includegraphics[height=1.5cm]{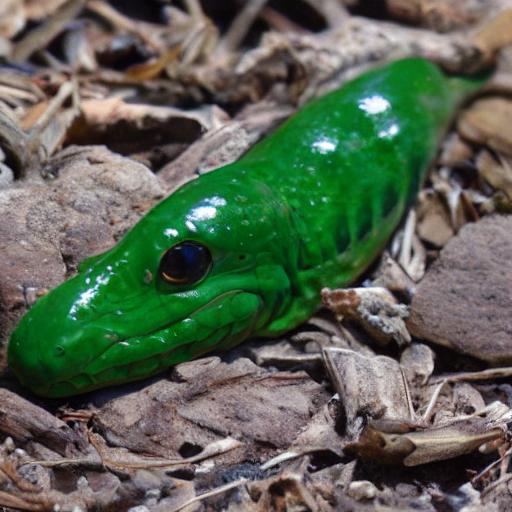}} &
    \raisebox{-0.45\height}{\includegraphics[height=1.5cm]{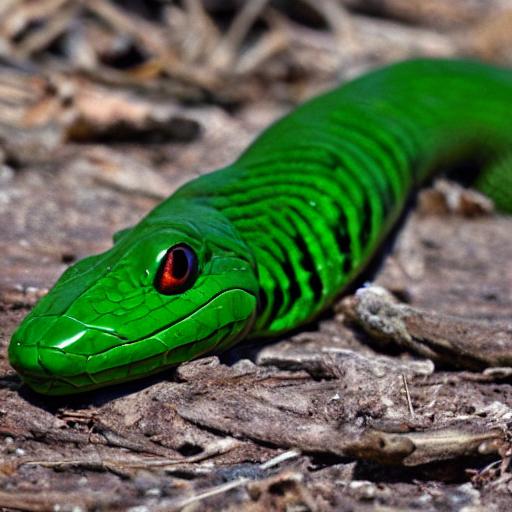}} &
    \raisebox{-0.45\height}{\includegraphics[height=1.5cm]{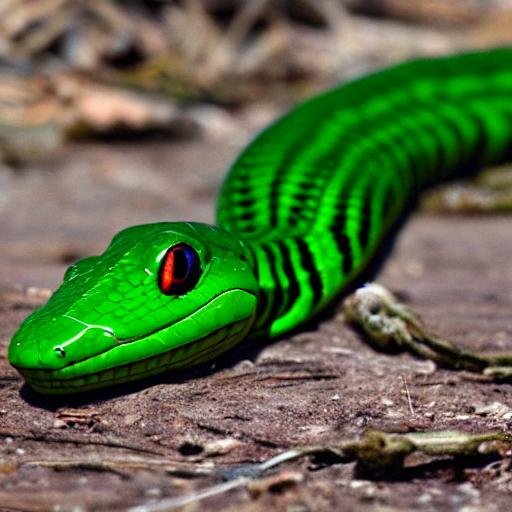}} \\
\end{tabular}
    \caption{Images for various \emph{clip\_guidance\_scale} ($cgs$) values. At $cgs=1.0$, the images are of  low quality. With $cgs$ set to higher values, the image sets lose diversity.}
    \label{fig:cgs}

\end{figure}

At $cgs=1.0$, images have no added compliance to the prompt but should be relatively ``likely'' with respect to SD's training set. In practice, however, the synthetic images appear largely incoherent.  %
Empirically, the coherence seems to be more consistent when $cgs \ge 2.5$.  Lower values produce images that are less compliant to the prompt, but lack of coherence makes them unsuitable for use in training.   In contrast, at $cgs=7.5$, images appear very well lit, highly saturated, and more  professionally composed.  Interestingly, this also makes them poor candidates for training as the ImageNet validation set \emph{does not} appear professional or unnaturally saturated.   Figure~\ref{fig:cgs} shows examples. For completeness, we trained using $cgs=1.0$ images (1,250,000 images created) and results were substantially lower than those in Table~\ref{table:resnet}\&\ref{table:resnet2}.

Consequently, we created a new set, \emph{Disamb2.5}, that uses the same prompts as Disamb, but uses ${cgs=2.5}$ (1,250,000 images created).  This significantly improved the performance, as shown in Table~\ref{table:resnet2}.

\begin{table}
  \small
  \begin{center}
  \begin{tabular}{c|c|c|c}

    & Training Set & Evaluation Set & Accuracy(\%)\\ %
    \hline
    \hline
    1. & ImageNet & Disamb2.5 & 59.2\\
    2. &ImageNet & PromptAug  &  50.9\\
    \hline
    3. &Disamb2.5 & ImageNet  &  43.4\\
    4. &Disamb2.5  & Disamb2.5  & 91.0  \\
    \hline
    5. &PromptAug & ImageNet  & 45.3 \\
    6.&PromptAug  & PromptAug & 81.2 \\

  \end{tabular}
   \caption{Accuracy of various train and evaluation sets, measured by training a ResNet-RS-152.}
 
  \label{table:resnet2}  
  \end{center}

\end{table}

The Disamb2.5 result in Table~\ref{table:resnet2}-line 3, indicates that a wider variety of ImageNet validation images are correctly classified when trained with this new dataset. Compared to the previous setting of $cgs=7.5$ for Disamb, this is a raw improvement of 17\%.   A byproduct of the increased diversity is that the synthetic images are more difficult for a standard, natural-image-trained network to classify (Table~\ref{table:resnet2}, line 1  vs. Table~\ref{table:resnet}, line 3).  With more diversity, the training and evaluation with Disamb2.5 (line 4) is harder than with Disamb $cgs=7.5$.

Next, rather than relying on \sd to \emph{implicitly} introduce diversity through reducing prompt adherence, we can also \emph{explicitly} encourage diversity through prompt augmentation. Similarly, \cite{radford2021learning} uses labels prefixed with ``a photo of a'', while \cite{pratt2022does} used prompts created with  GPT-3. Our next set, \textit{PromptAug}, creates a unique prompt for \textit{each} image -- even within a class. For this,  we used a combinatorial set of \{pre+post\}-fix prompt modifiers, and for each class selected 1,250.  See Figure~\ref{fig:promptAug}.

\begin{figure}
\small

\hrule height 1.5pt 
\smallskip
\noindent\textbf{Creation process for the \emph{PromptAug} Set}
\smallskip
\hrule height 0.1pt depth 0pt
\smallskip

$looks$ \random [beautiful, ugly, ..] \newline
$extent_{1,2}$ \random [slightly, very, ...] \newline
$typical$ \random [uncommon, typical, ...]\newline
$size$ \random [small, large, ...]\newline
$location$ \random [partially\_occluded, centered, ...]\newline
$style$ \random [overexposed, hyper\_sharp\_image, ...]\newline
\newline
prompt $\gets$ $looks + (extent_1 * typical) + (extent_2 * size) +\newline
\hspace*{0.55in} \mathbf {class\_name} + location  + style$

\caption{In \emph{PromptAug}, each image is generated with a unique prompt. 2 sample values shown for each modifier.   \random  randomly selects a single element from a set.  The full set is available to download. Samples are in Appendix~\ref{appendix:nuance}.}
\label{fig:promptAug}
\end{figure}

\begin{figure}[t]
\scriptsize
\setlength{\tabcolsep}{1.5pt}

\centering
\begin{tabular}{c|ccccc}
 & Synset & & Disamb & Disamb2.5 & PromptAug \\
\hline
\hline

\rotatebox[origin=c]{90}{\begin{minipage}{0.08\textwidth}\centering
jay\end{minipage}} & 
\raisebox{-0.45\height}{\includegraphics[width=0.08\textwidth]{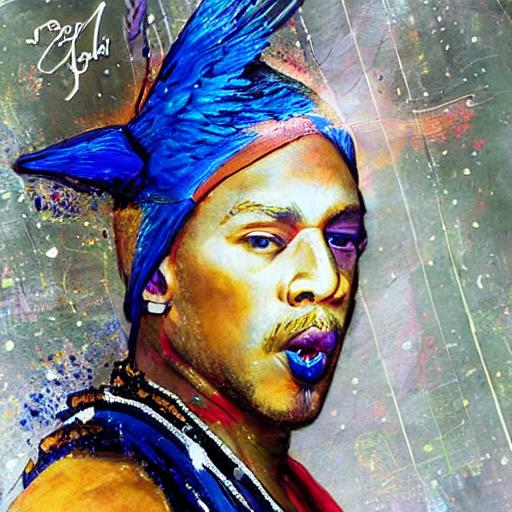}} &
\rotatebox[origin=c]{90}{\begin{minipage}{0.08\textwidth}\centering
a bird of\\type jay\end{minipage}} & 
\raisebox{-0.45\height}{\includegraphics[width=0.08\textwidth]{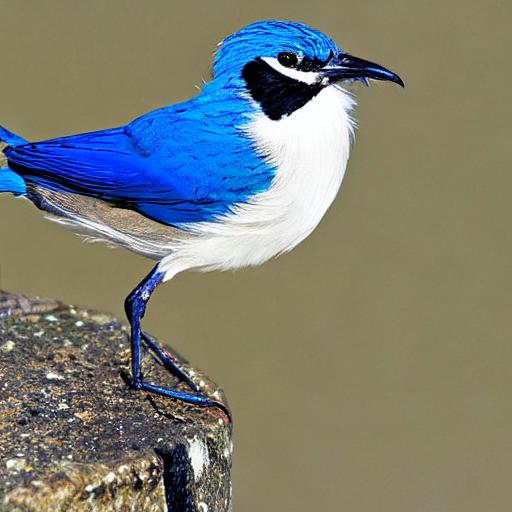}} &
\raisebox{-0.45\height}{\includegraphics[width=0.08\textwidth]{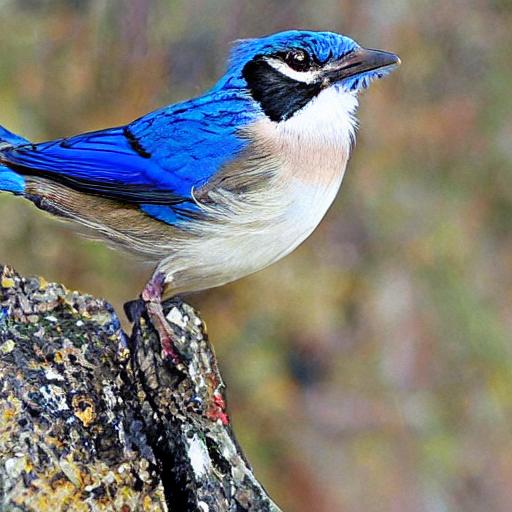}} &
\raisebox{-0.45\height}{\includegraphics[width=0.08\textwidth]{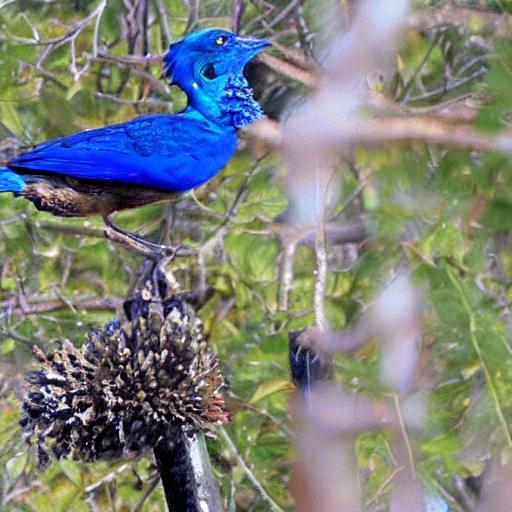}} \\

\rotatebox[origin=c]{90}{\begin{minipage}{0.08\textwidth}\centering
breastplate\\aegis,egis\end{minipage}} & 
\raisebox{-0.45\height}{\includegraphics[width=0.08\textwidth]{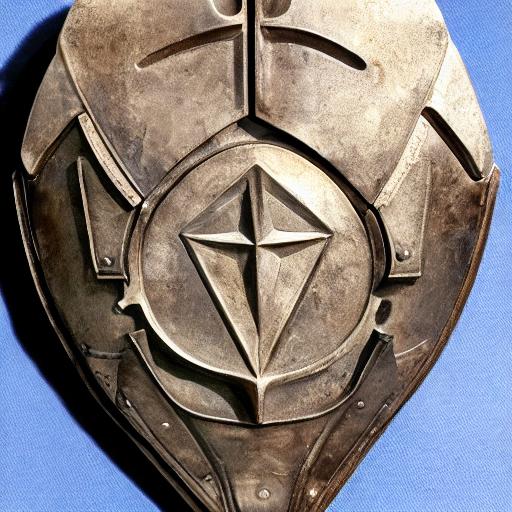}} &
\rotatebox[origin=c]{90}{\begin{minipage}{0.08\textwidth}\centering
breastplate\end{minipage}} & 
\raisebox{-0.45\height}{\includegraphics[width=0.08\textwidth]{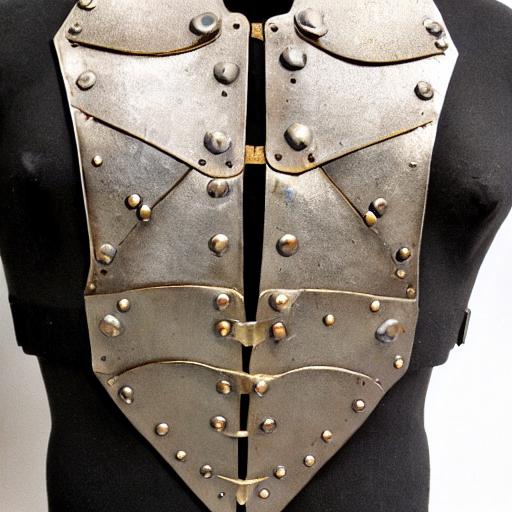}} &
\raisebox{-0.45\height}{\includegraphics[width=0.08\textwidth]{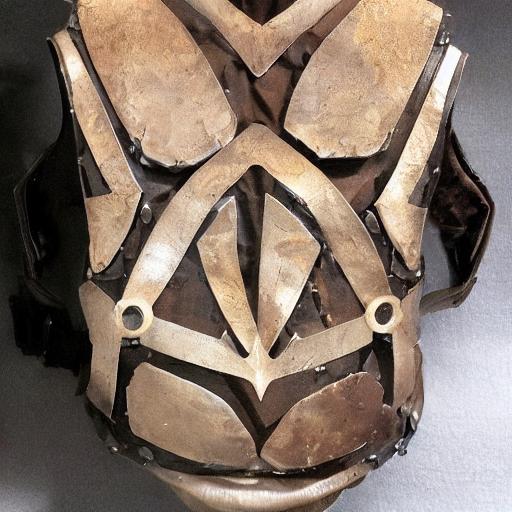}} &
\raisebox{-0.45\height}{\includegraphics[width=0.08\textwidth]{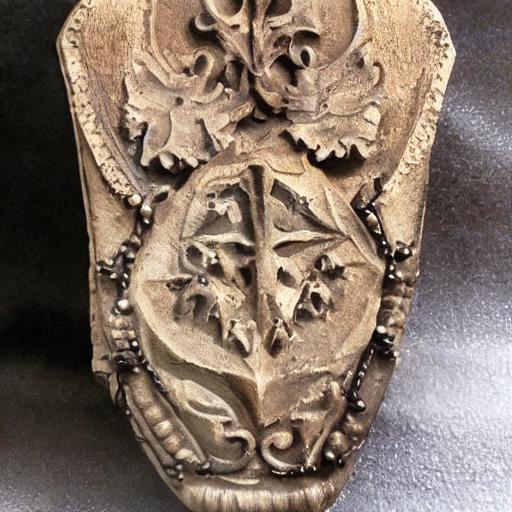}} \\

\rotatebox[origin=c]{90}{\begin{minipage}{0.08\textwidth}\centering
mortar\end{minipage}} & 
\raisebox{-0.45\height}{\includegraphics[width=0.08\textwidth]{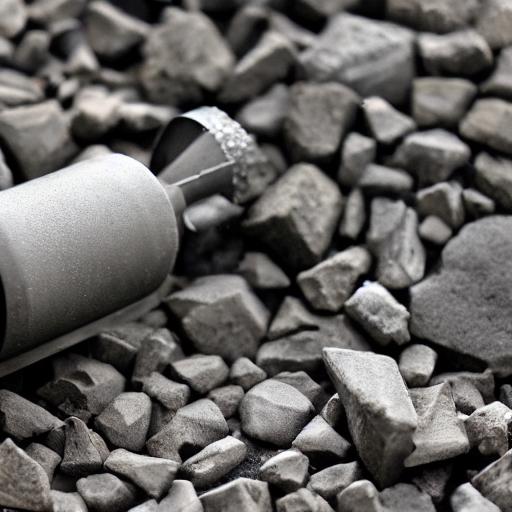}} &
\rotatebox[origin=c]{90}{\begin{minipage}{0.08\textwidth}\centering
mortar and\\pestle\end{minipage}} & 
\raisebox{-0.45\height}{\includegraphics[width=0.08\textwidth]{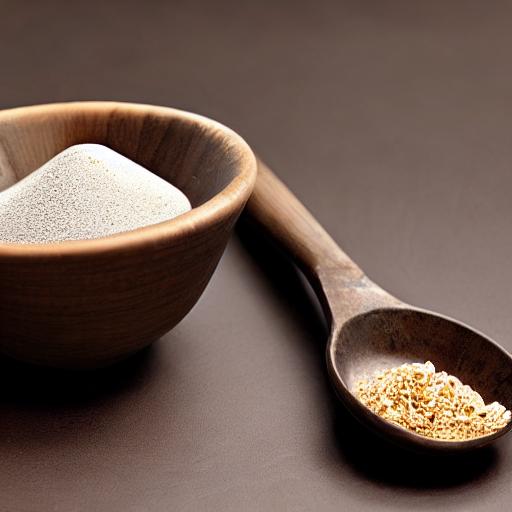}} &
\raisebox{-0.45\height}{\includegraphics[width=0.08\textwidth]{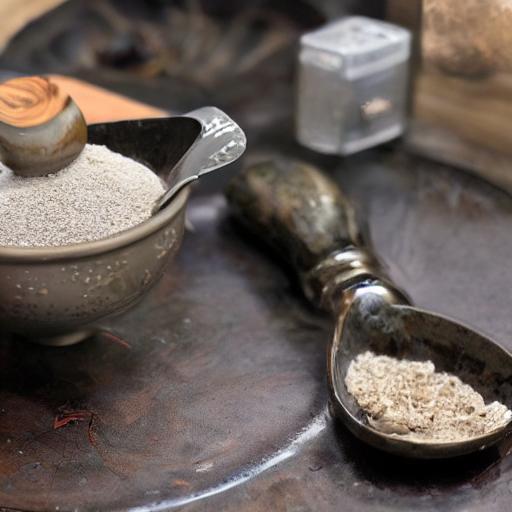}} &
\raisebox{-0.45\height}{\includegraphics[width=0.08\textwidth]{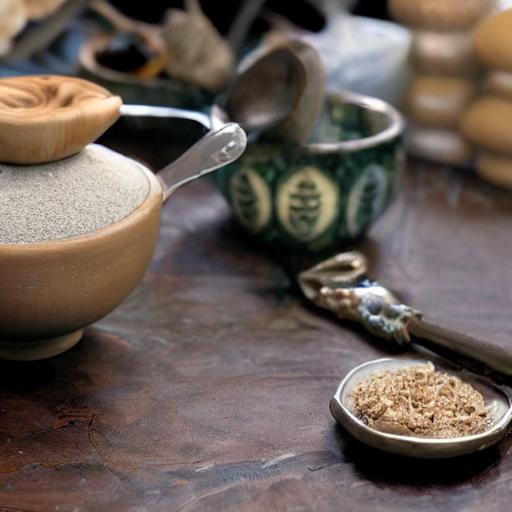}} \\

\rotatebox[origin=c]{90}{\begin{minipage}{0.08\textwidth}\centering
crane\end{minipage}} & 
\raisebox{-0.45\height}{\includegraphics[width=0.08\textwidth]{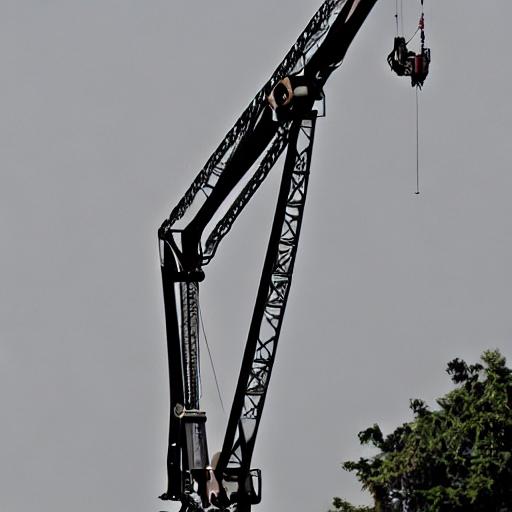}} &
\rotatebox[origin=c]{90}{\begin{minipage}{0.08\textwidth}\centering
a crane\\bird\end{minipage}} & 
\raisebox{-0.45\height}{\includegraphics[width=0.08\textwidth]{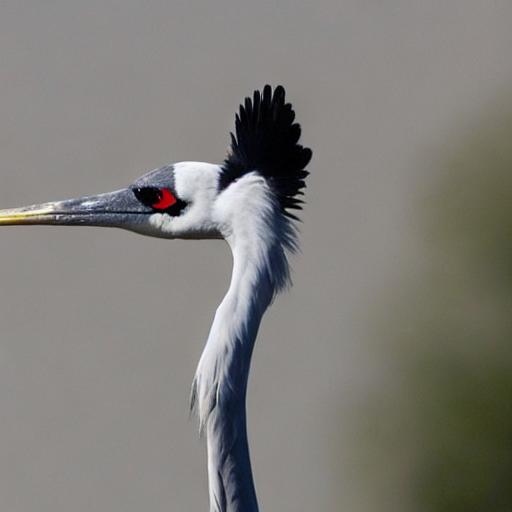}} & 
\raisebox{-0.45\height}{\includegraphics[width=0.08\textwidth]{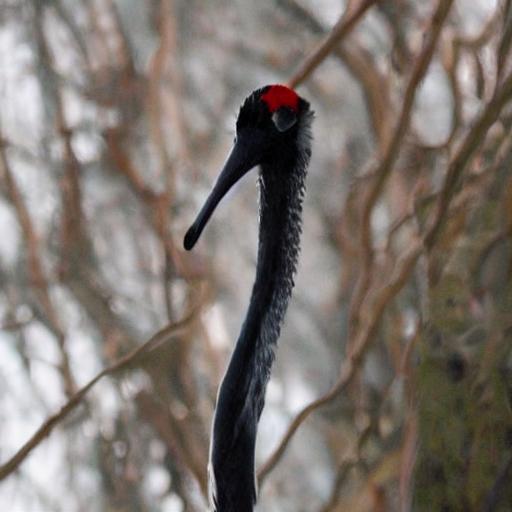}} & 
\raisebox{-0.45\height}{\includegraphics[width=0.08\textwidth]{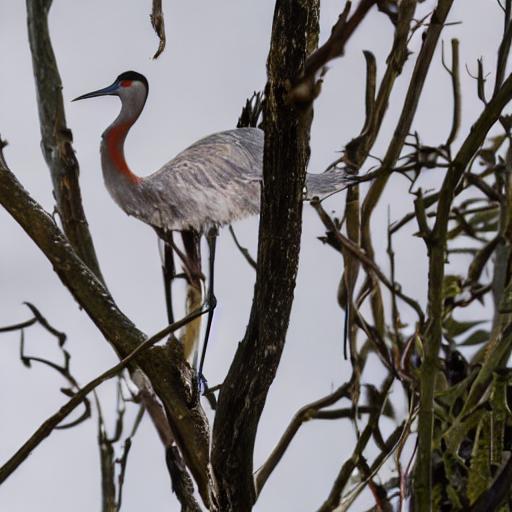}} \\

Centroid\\Distance & n/a & & $1.23{\times}10^{-5}$ & $2.05{\times}10^{-5}$ & $2.49{\times}10^{-5}$ \\
\end{tabular}
  \caption{Four classes, their prompts, and an image from each class,
  all using the same initial diffusion noise (\eg same seed). All four classes were  in the set of 105 prompts modified for Disamb. The \cd increases substantially with each set (n/a for Synset as the prompts are ambiguous).  PromptAug prompts are created by the procedure in Figure~\ref{fig:promptAug} using the shown class names. }
  \label{fig:set-samples}    
\end{figure}

Each modifier takes on a single, randomly chosen, value from its small set.  The modifiers are broadly applicable and can be used with any class label.   No adjectives describing color, texture, or feel were used as they inevitably would be inappropriate for some objects.    As can be seen in Table~\ref{table:resnet2}, using the images of this set yields a small improvement in accuracy (line 5 vs. line 3). Again, if we look at the indicators of greater diversity in these synthetic images, when ImageNet-trained networks are used for classification,  the accuracy drops (Table~\ref{table:resnet2}, line 2 vs. line 1).   When PromptAug images are used for both training and evaluation, accuracy is again reduced in comparison to previous experiments (line 6 vs. line 4). Figure~\ref{fig:set-samples} presents synthesized images in  each set.

\begin{figure}
    \centering
    \includegraphics[width=\linewidth]{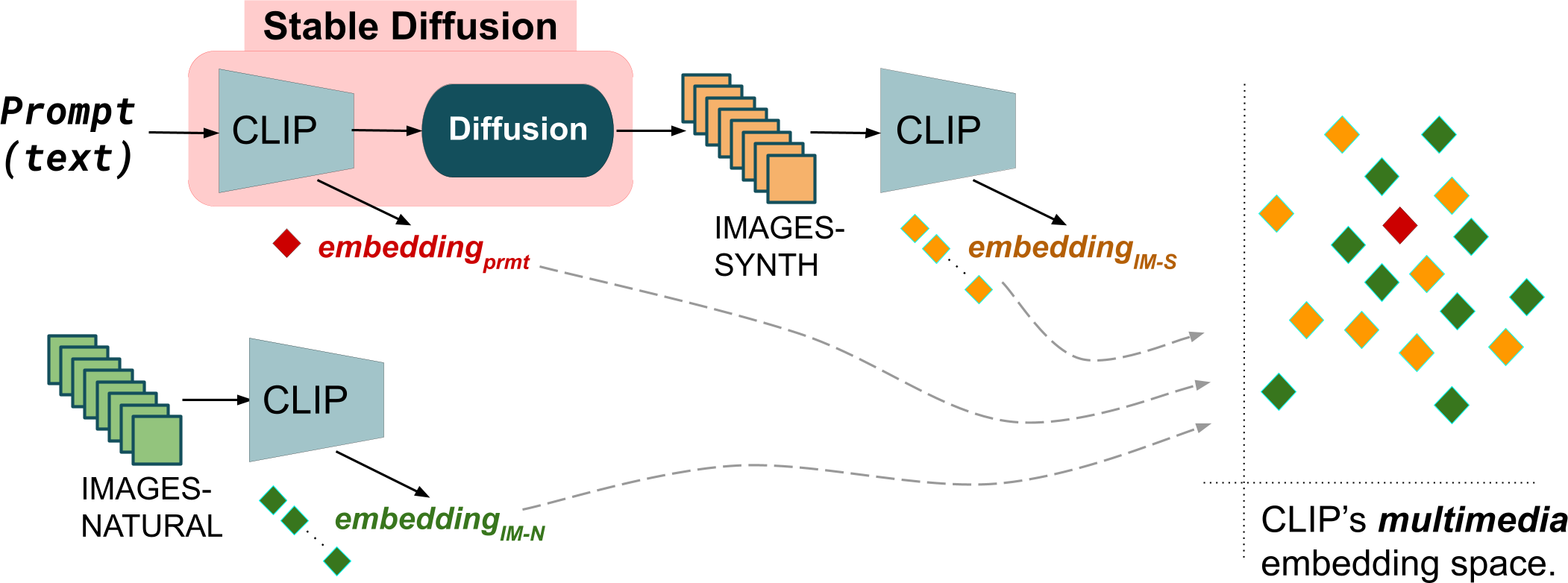}
    \caption{The prompt is passed to CLIP which yields an $embedding_{prmt}$ and guidance for SD. SD yields synthetic images (IM-S), which are passed back into CLIP to get $embeddings_{IM-S}$.   Natural images (IM-N) are passed to CLIP which returns $embeddings_{IM-N}$.  Right: All the embeddings (prompt, IM-N, IM-S) in the same, joint, embedding space.}
    \label{fig:procedure}
\end{figure}

The accuracy results are \emph{an effect of diversity}; they are not, in themselves, a measure of diversity.  
Naive methods of directly measuring diversity of images by examining pixels intensities, or even summary stats such as eigenvectors (PCA, \emph{etc.}) do not capture semantic content. Thankfully, modern language/image contrastive methods can provide a more meaningful measure. The CLIP model \cite{radford2021learning} is trained to create a joint high dimensional representation of both text and images; see Figure~\ref{fig:procedure}.

We utilize distances in the CLIP embeddings space to quantify diversity by computing centroids and summaries of image distances from them.\footnote{To be concrete, these are the 768 real-value CLIP multimedia embeddings.}
For all of the images, we compute their CLIP embeddings, $E_{S,c} = ClipEmbedImage(I_{S,c})$ where $I_{S,c}$ are the images of set $S$ in class $c$.  Because CLIP is trained using cosine similarity and relies on unit vectors, to calculate the centroid location of $E_{S,c}$, $M_{S,c} = uvec(\sum_i uvec(e_{S,c}^i \in E_{S,c}))$ where $uvec(\vec{v}) = \frac{\vec{v}}{|\vec{v}|}$ is the unit vector function.   Given the centroid, we then measure the \textit{Centroid Distance}, the mean squared distance from that centroid to each image embedding, for each class $Distance_{S,c} = \sum_i \codiff (e_{S,c}^i \in E_{S,c}, M_{S,c})^2 / |E_{S,c}|$ where $\codiff$ is the complement of cosine similarity, $\codiff(x, y) = 1 - uvec(x) \cdot uvec(y)$, and for each set by averaging across classes.

The \cd for each set is shown in Figure~\ref{fig:set-samples}. 
As expected, the \cd increases (gets more diverse), 
$Disamb$    $(1.23{\times}10^{-5}) <$  
$Disamb2.5$ $(2.05{\times}10^{-5}) <$ 
$PromptAug$ $(2.49{\times}10^{-5})$.
In comparison, ImageNet has greater diversity than all: $2.98{\times}10^{-5}$.  This provides insight into the distribution of the samples and why some natural images are incorrectly classified when using the synthetic examples for training --  the diversity of samples needed  is simply not present in the synthetic sets.

We've seen that increasing diversity in synthetic training sets, either implicitly or explicitly, causes increased accuracy and that diversity can be measured using Centroid Distance. However, introducing too much diversity  can  lower accuracy. We will examine this  in Reasons \#3 \& \#4.

For comprehensiveness, we investigated two alternatives to Centroid Distance. First, to test if another embedding would work better, all of these results were recreated with Inception embeddings \cite{borji2022}; these are the embeddings derived from the penultimate layers of Inception networks trained to classify ImageNet.  Inception embeddings did not provide any benefit to using CLIP embeddings and were not as applicable to non-ImageNet classes as the CLIP embeddings.  Further, CLIP maps prompts to the same embedding space as images, a key feature that will be used in the next section.  

Second, we applied the commonly used Frechet Distance (FD) $d^2 = |\mu_X - \mu_Y| + tr(\sum_X + \sum_Y - 2(\sum_X\sum_Y)^{1/2})$.  Instead of using FID Inception scores \cite{fid}, we substituted the CLIP embeddings. The first term of FD, subtracting centroids, strongly correlated with accuracy, but the second term, which compares across channels, did not.  \cite{chong2020effectively} also found limitations in FID. As such, we  use \cd going forward, which is also  based on distances from centroids.

\subsection{Reason \#3: Diversity in Diffusion}
\label{sec:variance-in-diffusion}
In this section, we separate the effects of \meanshift, measured as the difference in set centroids $\codiff(r_1, r_2) = 1 - uvec(r_1) \cdot uvec(r_2)$, and diversity on accuracy.  We will also isolate the contribution of diffusion by expanding the use of \cd from the previous section. 
Recall that CLIP maps images and text prompts into the same space.   Zero-shot classifiers using CLIP \cite{radford2021learning} find the nearest neighbor of an image embedding, the \textit{query}, among the set of embeddings of the class prompts, the \textit{references} (one for each of the 1000 classes).

Generalizing this, we allow both query and reference roles to be either prompts or images.  Additionally, both prompts and images can have multiple instances per class. When the reference set has multiple instances ($N > 1$) in a class, we replace those $N$ embeddings with their single centroid. The accuracy calculation is the same: a query is correct if its label matches the label of the nearest centroid's class.  Analagously to Centroid Distance,  we'll refer to this as \textit{Centroid Accuracy}; see Figure~\ref{fig:clusters}. Formal notation is in Appendix~\ref{appendix:centroid-accuracy}.

\begin{figure}
\centering
 \includegraphics[height=1.45in]{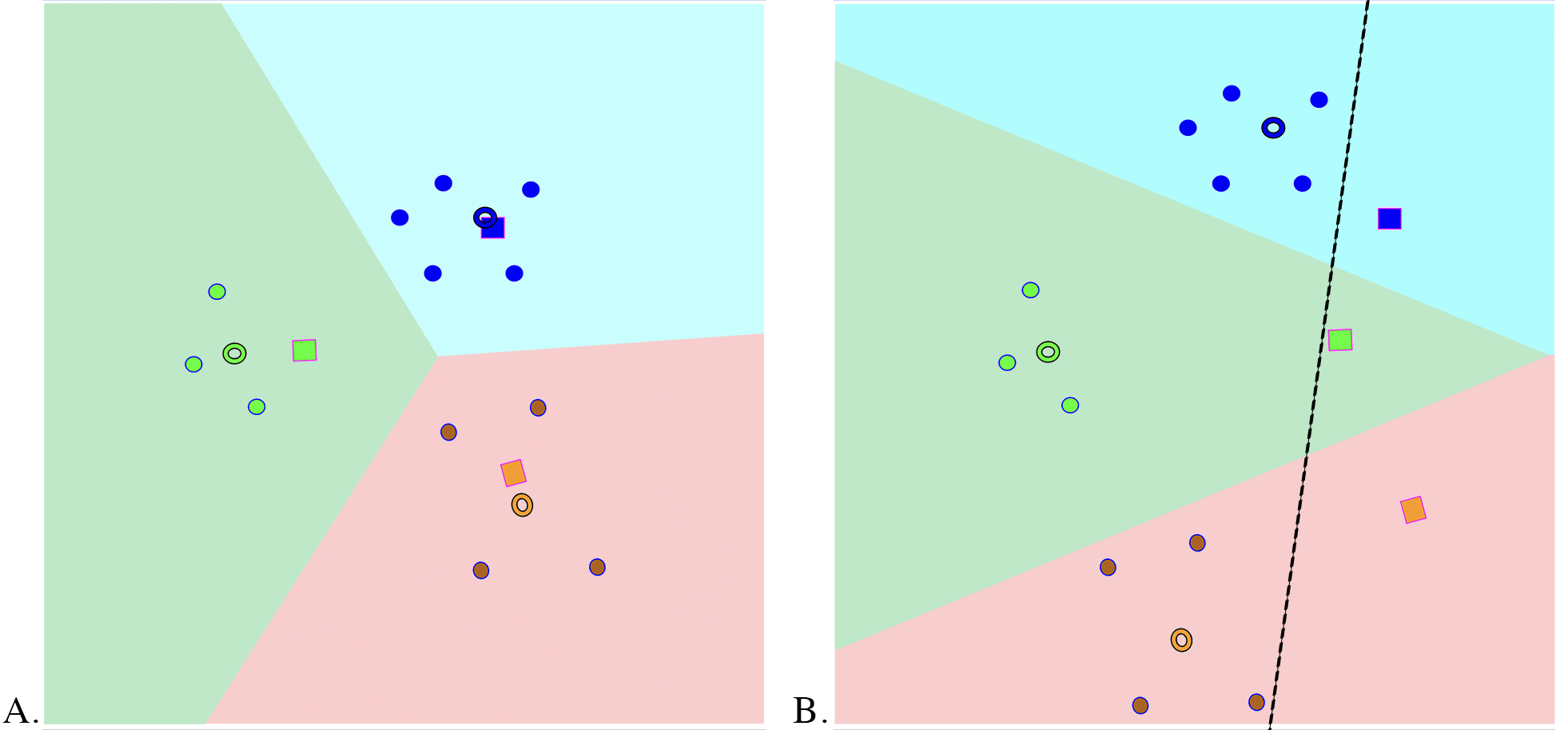}
 
  \caption{3 clusters in the embedding space.  Circles: images; donuts: centroid of the images;  squares: text  prompt.  Left: Typical clustering. Red: images have higher diversity, but close centroid to the prompt.  Blue: smaller distribution, and perfect centroid.  Green: small diversity, but centroid is far from generating prompt (flawed, but similar images).   Right: An alternate view, same matches, but all the prompts are \emph{linearly separated} from all the images (discussed in Section~\ref{sec:modified-prompts}).}
  \label{fig:clusters}    

\end{figure}

We compute Centroid Accuracy on both images and prompts, distinguished with a -IM and -PRMT suffix respectively.  Experiments include using the image-sets: ImageNet-IM, Disamb-IM, Disamb2.5-IM, PromptAug-IM,  and the prompt-sets: Disamb2.5-PRMT, PromptAug-PRMT, as both query and reference.   The full set of 36 classification trials is given in  Appendix~\ref{appendix:centroid-accuracy}; we will show the more informative ones as they are discussed.   Whichever role is used (reference or query), references always use the set's training split, and queries always uses an independent evaluation split.  

In Table~\ref{table:vid-centroid-accuracy}, Experiment 5 is the baseline, and is comparable to standard zero-shot learning where prompts are the reference set and the ImageNet-validation images are  queries.  The Centroid Accuracy for ImageNet (measured by the closest prompt to the image in the embedding space) is 70.1\%.  

Experiment 17 repeats the above experiment with  Disamb2.5-IM as the queries to yield 74.9\% accuracy. The 25.1\% error can be decomposed into the portions caused by \meanshift\space and by diversity.  To ascertain the relative impact of each, another experiment is performed.  Experiment 15 uses Disamb2.5-IM's \emph{own} centroid as the reference set rather than the prompts.  By calculating the Centroid Accuracy using the same distributions of reference and query, only errors due to diversity remain. If diffusion created images with little/no diversity, then every image would map back to the centroid and Centroid Accuracy would be 100\%.   However, accuracy is 82.0\% (18.0\% error), indicating diffusion's diversity is large enough that some images are closer to the centroids of other classes. Comparing this to the 25.1\% error when using prompts as references, because only the centroid has changed between experiments, the difference in the errors is attributable to a \meanshift.  A \meanshift\space within the CLIP embedding space is a shift in the concept that is represented -- a \textit{meaning-shift}.

\begin{table}

\setlength{\tabcolsep}{2pt}
\small
    \centering
    \begin{tabular}{c|c|c|c|c}
     Reference & Query   & Centroid & Avg Cos. & Exp  \\
          Set & Set     & Accuracy & Similarity & \# \\
     \hline
     \hline
     Disamb2.5-PRMT & ImageNet-IM & 70.1\% & 0.2425 & 5 \\
     Disamb2.5-PRMT & Disamb2.5-IM & 74.9\% & 0.2421 & 17\\
     Disamb2.5-IM & Disamb2.5-IM  & 82.0\% & 0.8709 & 15 \\

    \end{tabular}
    \caption{Centroid Accuracies.  Experiment \#'s index to the full set of 36 \{reference$\times$query\} tests. Full list in  Appendix~\ref{appendix:centroid-accuracy}.}
    \label{table:vid-centroid-accuracy}
    \vspace{-0.15in}
\end{table}

Returning to Experiment 15, a second interesting aspect is that we can attribute the 18.0\% misclassification to diversity introduced by \emph{diffusion} rather than CLIP. We know this because the images are all created from the same prompt; therefore, diffusion is the only source of variation in the generative process.  This is an accuracy-based measure of the $Distance_{Disamb2.5} = 2.05{\times}10^{-5}$ from Figure~\ref{fig:set-samples}.

So far, in this section, to quantify the effects of \meanshift\space vs. diversity, we modified the role of the synthetic images to use them as the query set. We determined that both a shift between the centroids of the synthetic images and diversity of the synthetic images reduces Centroid Accuracy with the latter contributing more to the overall error.

This puts Reason \#2 in context.   In Reason \#2,  we needed to add diversity to improve the ResNet accuracies (Tables~\ref{table:resnet}\&\ref{table:resnet2}).  Here, we showed that just adding diversity is not a panacea; the images generated must still accurately represent the underlying prompt --- which the diffusion process does not guarantee.   This is why augmentations, such as in PromptAug, are used; they explicitly guide  diversity by providing hints to diffusion about how to create unique, but meaningful, images.  The augmentations did, on average, improve accuracy; useful diversity was added.   In a few classes, however, the errors from \meanshift\space dominated the performance and the prompt embedding from CLIP did not represent the class.  This failure mode is described in the next section.
 
On a pragmatic note, we observe that the diffusion process has varying levels of success in paying attention to all portions of specific prompts.  Because of the popularity of \tti systems, the practice of \emph{prompt tuning} \cite{hao2022optimizing, zhou2022learning, mjprompts} has become widespread; this is the colloquial name given to the explicit control of mean and diversity.  In Appendix~\ref{appendix:nuance}, we provide illustrative examples of prompts that do exactly as intended; they control diversity by narrowing the subject/background (the ``school bus'' class).  To contrast this, we also present a class for which diffusion has difficulty; it only creates a canonical version of the subject (``quail'').

\subsection{Reason \#4: Delving Deeper into Centroid-Shifts}
\label{sec:mean-shift}

To begin this section, recall that Experiment 5 in Table \ref{table:vid-centroid-accuracy} achieved a Centroid Accuracy of 70.1\%.  This performance indicates a reasonable match between the concept (text/class prompt) as encoded by CLIP and the ImageNet images (when also encoded by CLIP).   Diversity, as shown in the previous section, has a larger contribution to error than \meanshift.

\begin{figure}
\centering

\setlength{\tabcolsep}{1pt}
\setlength\extrarowheight{-10pt}
\begin{tabular}{ccccccc}

\rotatebox[origin=c]{90}{\begin{minipage}{0.08\textwidth}\centering
hammer
\end{minipage}} &    
\raisebox{-0.45\height}{\includegraphics[width=0.072\textwidth]{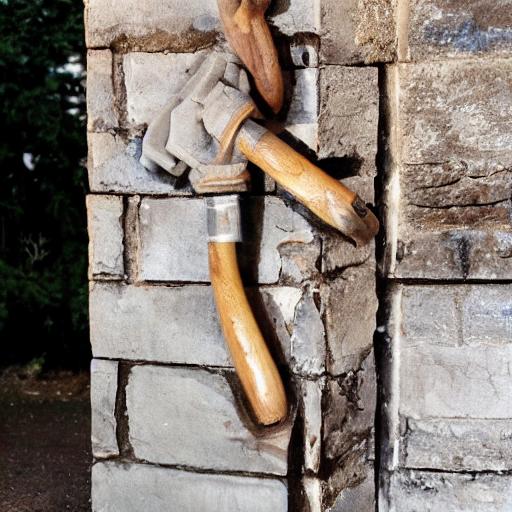}}&
\raisebox{-0.45\height}{\includegraphics[width=0.072\textwidth]{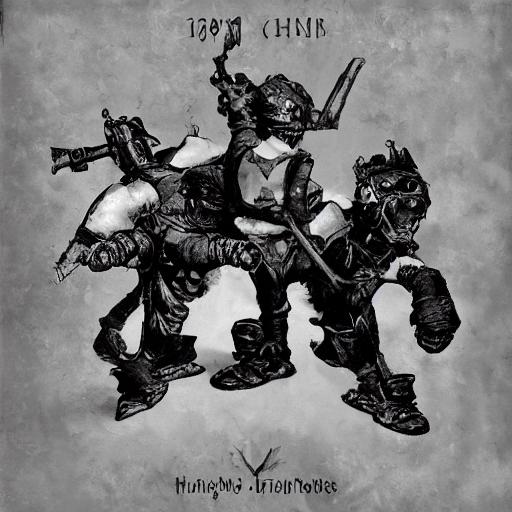}}&
\raisebox{-0.45\height}{\includegraphics[width=0.072\textwidth]{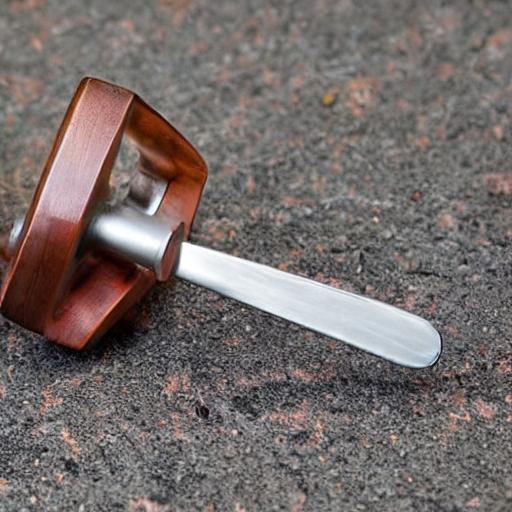}}&
\raisebox{-0.45\height}{\includegraphics[width=0.072\textwidth]{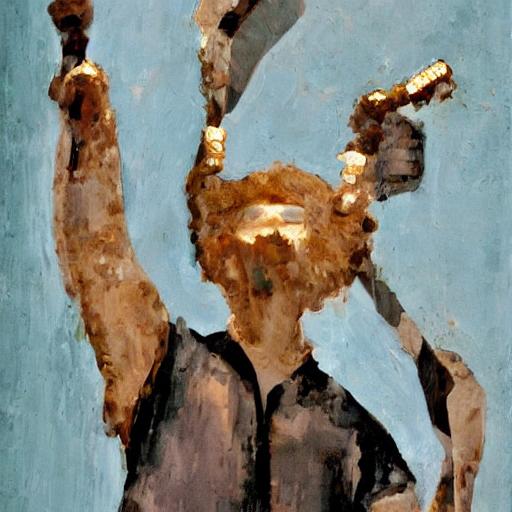}}&
\raisebox{-0.45\height}{\includegraphics[width=0.072\textwidth]{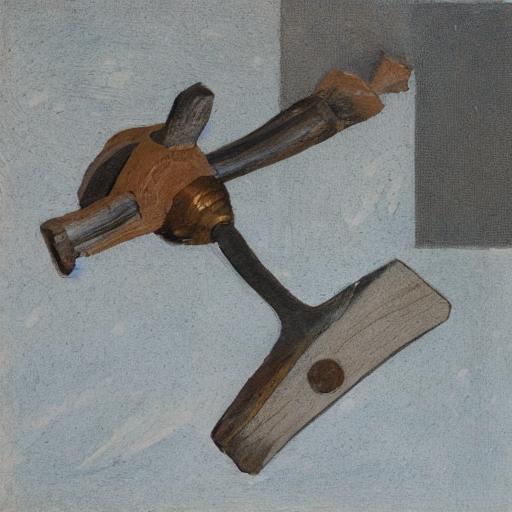}}& \\

\rotatebox[origin=c]{90}{\begin{minipage}{0.08\textwidth}\centering
nail
\end{minipage}} &    
\raisebox{-0.45\height}{\includegraphics[width=0.072\textwidth]{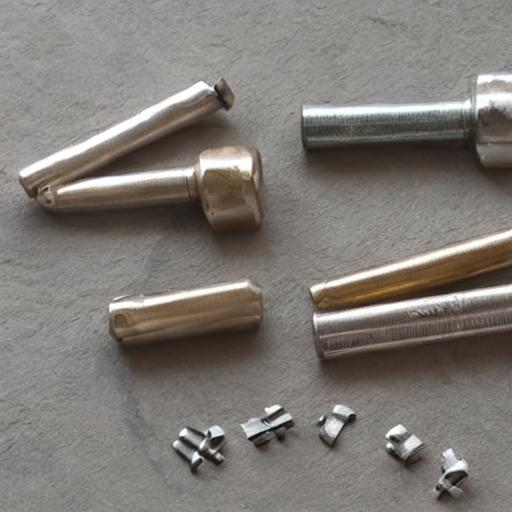}}&
\raisebox{-0.45\height}{\includegraphics[width=0.072\textwidth]{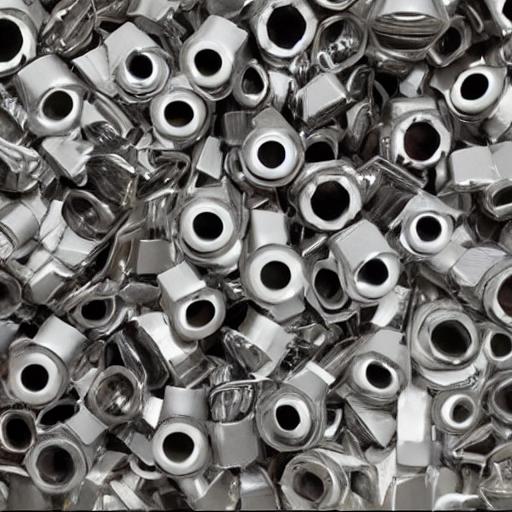}}&
\raisebox{-0.45\height}{\includegraphics[width=0.072\textwidth]{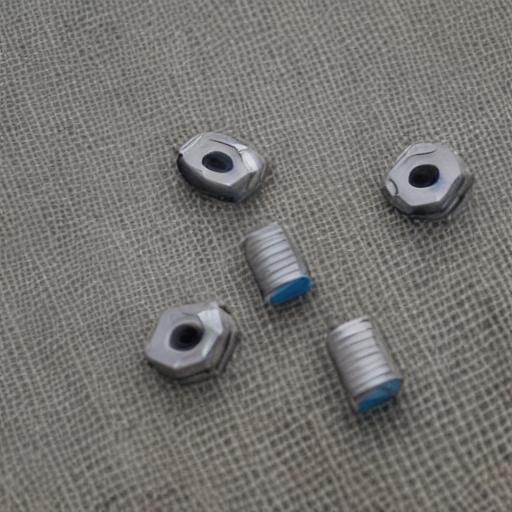}}&
\raisebox{-0.45\height}{\includegraphics[width=0.072\textwidth]{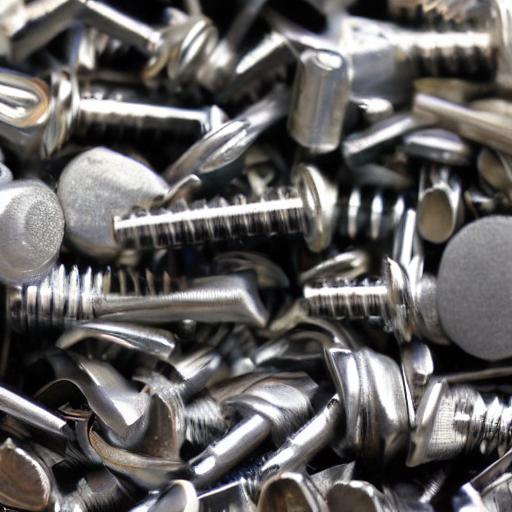}}&
\raisebox{-0.45\height}{\includegraphics[width=0.072\textwidth]{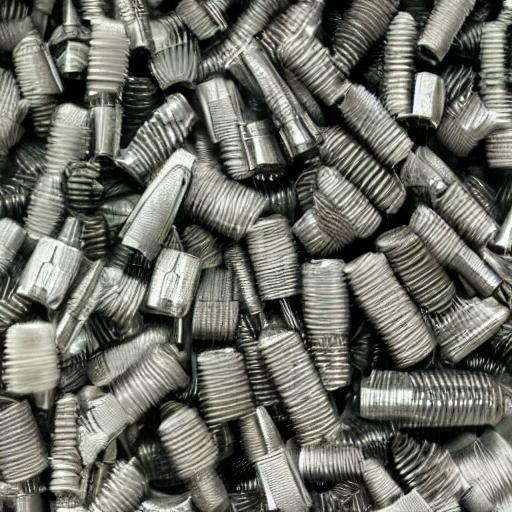}}

\end{tabular}
   \caption{Two example classes where the generated images are poor: "hammer" (top) and "nail" (bottom).}
  \label{fig:hammer}    

\end{figure}

Nonetheless, {\meanshift}s can be largely detrimental to some individual classes.  To see this effect, we examine classes that SD is  \textbf{unable to generate}. Two such classes are ``hammer'' and ``nail''; see Figure~\ref{fig:hammer}.   The synthetic images for ``hammer'' are largely nonsensical -- not resembling any coherent object.   In contrast, the synthetic images for ``nail'' represent real objects (screws, nuts and bolts), but not the desired class, ``nail.''  How do these two different failure modes manifest themselves in our analysis?  Table~\ref{table:classes} shows the accuracy for these two classes.

Looking at the poor results for hammer, we can ask whether diffusion's generation is to blame, or is the concept ``hammer'' incorrectly represented in CLIP?  To determine this, we set the reference set to Disamb2.5-PRMT (which includes ``hammer'') and first lookup the \emph{natural} ImageNet-IM images.  The accuracy is only 14\%  (Table~\ref{table:classes}, line 1) for this class, significantly lower than the average across all classes of 70.1\%.    Simply put, CLIP does not represent the prompt well.  The embedding CLIP produced, which should represent the concept ``hammer'', erroneously represents a different concept.  Additionally, we can ask whether diffusion is able to accurately recreate the concept that is represented.  Changing the query set to the synthesized images (line 2)  does not improve performance; the diversity from diffusion places images so far from their centroid that they are in a different class.

Next, we repeat the same test for  ``nail.''  As before, first we set the reference set to the Disamb2.5-PRMT and lookup the \emph{natural} ImageNet ``nail'' images.  As with hammers, the accuracy is lower than the average (58\%). Next, we ask whether  diffusion is able to accurately recreate whatever concept CLIP produced for ``nail.''  Interestingly, the accuracy jumps to 84\%.   Contrast this with the 16\% for ``hammer'' in the same test.  This indicates that, although the concept of ``nail'' may not be represented as accurately as other classes, the synthetic images generated by diffusion \emph{are} consistent with CLIP's flawed prompt embedding --- even though they are \emph{not} close to the ImageNet ``nail'' images.
This error is largely attributable to a \meanshift\space between the synthesized and ImageNet images.

In summary,``nail'' and ``hammer'' are failure  examples where the system would benefit from an improved CLIP model.   In fact, recently, \cite{wortsman2023ZeroShotAccuracy} employed an improved CLIP model that improves the zero-shot to 80.1\% (competitive with our  ResNet-RS-152 at 80.8\%; Table~\ref{table:resnet}).

\begin{table}
\small
  \setlength{\tabcolsep}{3pt}
  \begin{center}
  \begin{tabular}{c|c|c|c|c}

    Reference & Query & Avg.& Hammer & Nail \\
    Set & Set &CA  & CA  & CA \\
\hline
\hline
    Disamb2.5-PRMT & ImageNet-IM & 70.1  & 14.0  & 58.0\\
    Disamb2.5-PRMT & Disamb2.5-IM &  74.9  & 16.0  & 84.0\\

  \end{tabular}
   \caption{Centroid Accuracy (CA) results for  ``nail'' and ``hammer'' indicate different failure modes.  The Avg. CA column give the average accuracy across all 1000 classes.}

  \label{table:classes}  
  \end{center}
\end{table}

\section{Geometry of Embeddings}
\label{sec:modified-prompts}

In the previous section, we concentrated on Centroid Accuracy for classification. In this section, we take a closer look at the 
cosine similarity scores it is based on. We reveal some surprising findings in the underlying geometry of CLIP's shared embeddings space for prompts and images.

When we examine the magnitudes of the similarities between prompts and other prompts (Table~\ref{table:textPrompts}, Experiments 30,36,29) the similarity is high (0.70-0.90).  The same is true with the similarity of images (Experiments 22,15).  Now, let us examine the similarity \emph{across modalities}: between prompts and images.    There is a dramatic drop  in similarity without a correspondingly large drop in accuracy.

There seems to be a set of clusters that are prompts and another that are images.   Within the clusters, however, the individual classes are oriented such that the classification \emph{across} modalities is still possible.  We attempted to visualize this.   To do so, we first wanted to examine if there was a clean separation between the set of prompt embeddings and the set of image embeddings.   To our surprise,    there was not only a separation, but the two types of embeddings were \emph{linearly} separable.  A potential way to have them linearly separable, while still useful for classification, is shown in Figure~\ref{fig:clusters}B.  

\begin{table}
\small
\setlength{\tabcolsep}{1.5pt}
    \centering
    \begin{tabular}{c|c|c|c|c}
       &                         & Centroid   & Avg Cos.  & Exp  \\
      Reference Set & Query Set  & Accu- & Simi- & \# \\
      &                          & racy   & larity  & \\
     \hline
     \hline
     PromptAug-PRMT & PromptAug-PRMT  & 100\% & 0.878& 30 \\ 
     PromptAug-PRMT & Disamb2.5-PRMT & 99.6\% & 0.796& 36 \\       
     Disamb2.5-PRMT & PromptAug-PRMT & 98.8\% & 0.699& 29 \\ 
     \hline
     PromptAug-IM & PromptAug-IM  & 70.5\% & 0.852 & 22 \\
     Disamb2.5-IM & Disamb2.5-IM  & 82.0\% & 0.871 & 15 \\          
\hline
     Disamb2.5-PRMT & Disamb2.5-IM  & 74.9\% & 0.242 &  17\\
     Disamb2.5-PRMT & PromptAug-IM  & 62.4\% & 0.231 &  23\\ 
     PromptAug-PRMT & Disamb2.5-IM & 76.2\% & 0.258&  18\\ 
     PromptAug-PRMT & PromptAug-IM  & 69.2\% & 0.261&  24\\ 
     
    \end{tabular}
    \caption{%
    Measures within and across text \& image modalities.}%
    \label{table:textPrompts}
\end{table}

Examining CLIP's loss function provides insight into how this occurs.  The loss function rewards  correct pairings of prompt and image, but does not attempt to create the same representation for the pair.  This explains how this result is possible, though it remains quite unexpected.  Recently, \cite{linear} have also discovered this.

\section{\knn vs Centroid Accuracy}
\label{sec:knn}

Zero-shot classification typically employs a single point, the prompt embedding of each class label, as references vectors.  This effectively creates a \knn classifier  with $k=1$.  Directly relating to our discussions of class diversity, the collapsing of the reference set to a single point may mask the shape and extent of the full reference set.   If we expand to the general \knn approach,  this potential loss of information is no longer necessary.  This is especially relevant to the scenarios where   the reference set is comprised of labeled images or where unique prompts (PromptAug) are used.
We repeat the entire set of experiments \textbf{without collapsing the reference set to centroids}, and instead find the nearest (highest cosine similarity) $k$ reference set embeddings.  The most frequent class of the found $k$ is returned as the label. We test $k=1,5$.

Though we do not have the space to delve into the details of the 60 \knn experiments, the full  results are in  Table~\ref{table:centroid-accuracy} (Appendix~\ref{appendix:centroid-accuracy}).  Across the numerous combinations of reference and query sets, Centroid Accuracy averages to 77.1\% while \knn $k=1$ accuracy is 74.6\% and $k=5$ is 77.3\%. $k=5$ was better than Centroid Accuracy in 40\% of the experiments.

This is consistent with the findings presented so far.  Without averaging, when using the single nearest neighbor, if the reference set produces imperfect examples, performance can degrade.  However, when the labels of a few samples are considered,  the effect is mitigated. The results indicate that here, reference centroids sufficiently summarized the statistic required for classification.
Additionally, on a pragmatic note,  Centroid Accuracy performs similarly to $k=5$ and it is computationally far less expensive (needing to consider 1000 centroids vs. 1.2M neighboring images).  In the future, it remains an open question of whether more sophisticated methods beyond \knn can be used to better exploit the distributions of the individual examples.

\section {Discussion}

The last few years have seen an explosion in both the number of \tti systems and their improvement in quality. Rapid innovation has taken place in all aspects of the systems: the underlying large language model, the generation process, the memory requirements and the computational speed.  Despite, or perhaps because of, the rate of innovation coupled with the already well-understood mathematical underpinning of the diffusion process, comparatively little work has been devoted to developing an understanding about what can be generated.  Though the synthetic images already often appear indistinguishable from natural ones, they are not suitable for training classifiers.  This indicates that the \emph{set} of generated images does not yet represent the \emph{set} of natural images.    We examined the roles of diversity and centroid-shifts in both the CLIP and diffusion processes.  Our findings include:

\begin{enumerate}
    \item Ambiguity in prompt interpretation hinders training with synthetic images. It can be mitigated by revising prompts to better reflect the class semantics. This is not merely a matter of synonyms and homonyms -- the target class may itself have a bias that is not represented in the prompt.

    \item %
    The influence of the prompt must be carefully controlled. 
    Allowing a high influence does not allow for sufficient diversity when creating new datasets (Table~\ref{table:resnet}). Reducing the influence increases diversity (Figure~\ref{fig:set-samples}), which can improve their utility in training (Table~\ref{table:resnet2}).  However,   increasing diversity has limits. Coherence and accuracy of the images can drop as diversity increases.  As demonstrated, a few classes  increased  diversity too much -- \eg to the point where they are no longer accurate (Table~\ref{table:classes}) and are, therefore, no longer useful in training.

    \item Accuracy is affected by both \meanshift\space and diversity. Our Centroid Accuracy and \cd techniques in combination isolate the effects of diversity and show that diversity has a larger impact on accuracy than \meanshift. In addition, the effects of the CLIP and diffusion processes can be measured separately. Diffusion is a larger contributor to error than CLIP.

    \item For most classes, CLIP faithfully reproduces the concepts; it largely represents  ImageNet's interpretation.  Nonetheless, a potential failure case for any \tti system is its inability to recreate images for some prompts/concepts. We presented examples where CLIP represented  nonsensical concepts as well as incorrect  concepts.

    \item In terms of CLIP's representation of text and images, we have found that the embeddings are \emph{linearly separable} --- an intuitively  surprising finding as successful zero-shot classifiers and most clustering is rarely visualized as such.  This explains why prompts that are internally consistent when embedded do not always translate to internally consistent images when used by diffusion. 
\end{enumerate}

Our study has used Stable Diffusion, the most commonly used \tti system; however, any generation system can be easily substituted.   We hope that this work provides insights into image generation systems and also provides guidelines for the understanding, analysis, and evaluation of new systems as they are rapidly deployed.

\appendix

\clearpage
\section{Appendix: Centroid Accuracies and \knn}
\label{appendix:centroid-accuracy}

\begin{table}[!b]
\begin{minipage}{\textwidth}
\small
  \begin{center}
    \begin{tabular}{|c|l|l|c|c|c|c|}
      \hline
Experiment & Reference  &Query                    & Centroid & \knn k=1  & \knn k=5  & Avg Cos. \\
\#  &  Set & Set                    & Accuracy & Accuracy & Accuracy & Similarity \\      

\hline
1   & ImageNet-IM&ImageNet-IM                 & 75.1\% & 76.3\% & 79.2\% & 0.8353 \\
2   & Disamb-IM&ImageNet-IM                       & 60.8\% & 60.9\% & 63.1\% & 0.7748 \\
3   & Disamb2.5-IM&ImageNet-IM                       & 61.2\% & 59.4\% & 62.5\% & 0.7779 \\
4   & PromptAug-IM&ImageNet-IM                       & 61.4\% & 58.3\% & 62.0\% & 0.7762 \\
5   & Disamb2.5-PRMT&ImageNet-IM              & 70.1\% & 70.1\% & 70.1\% & 0.2425 \\
6   & PromptAug-PRMT&ImageNet-IM              & 73.9\% & 72.4\% & 72.8\% & 0.2600 \\
7   & ImageNet-IM&Disamb-IM                       & 76.6\% & 71.2\% & 75.3\% & 0.8388 \\
8   & Disamb-IM&Disamb-IM                             & 82.0\% & 95.5\% & 96.3\% & 0.9685 \\
9   & Disamb2.5-IM&Disamb-IM                             & 89.8\% & 89.5\% & 91.3\% & 0.8869 \\
10  & PromptAug-IM&Disamb-IM                             & 82.5\% & 82.1\% & 84.9\% & 0.8658 \\
11  & Disamb2.5-PRMT&Disamb-IM                    & 86.3\% & 86.3\% & 86.3\% & 0.2560 \\
12  & PromptAug-PRMT&Disamb-IM                    & 86.2\% & 85.3\% & 85.4\% & 0.2675 \\
13  & ImageNet-IM&Disamb2.5-IM                       & 66.0\% & 58.8\% & 63.5\% & 0.8072 \\
14  & Disamb-IM&Disamb2.5-IM                             & 78.1\% & 81.1\% & 82.8\% & 0.8518 \\
15  & Disamb2.5-IM&Disamb2.5-IM                             & 82.0\% & 80.8\% & 83.9\% & 0.8709 \\
16  & PromptAug-IM&Disamb2.5-IM                             & 76.7\% & 72.7\% & 76.9\% & 0.8564 \\
17  & Disamb2.5-PRMT&Disamb2.5-IM                    & 74.9\% & 74.9\% & 74.9\% & 0.2421 \\
18  & PromptAug-PRMT&Disamb2.5-IM                    & 76.2\% & 74.2\% & 74.5\% & 0.2580 \\
19  & ImageNet-IM&PromptAug-IM                       & 55.8\% & 49.3\% & 53.7\% & 0.7874 \\
20  & Disamb-IM&PromptAug-IM                             & 59.2\% & 60.5\% & 62.6\% & 0.8136 \\
21  & Disamb2.5-IM&PromptAug-IM                             & 61.7\% & 61.2\% & 64.3\% & 0.8375 \\
22  & PromptAug-IM&PromptAug-IM                             & 70.5\% & 65.6\% & 70.3\% & 0.8518 \\
23  & Disamb2.5-PRMT&PromptAug-IM                    & 62.4\% & 62.4\% & 62.4\% & 0.2314 \\
24  & PromptAug-PRMT&PromptAug-IM                    & 69.2\% & 67.5\% & 67.9\% & 0.2605 \\
25  & ImageNet-IM&PromptAug-PRMT              & 93.1\% & 80.1\% & 88.0\% & 0.2742 \\
26  & Disamb-IM&PromptAug-PRMT                    & 92.9\% & 83.1\% & 88.0\% & 0.2596 \\
27  & Disamb2.5-IM&PromptAug-PRMT                    & 93.0\% & 80.4\% & 87.1\% & 0.2596 \\
28  & PromptAug-IM&PromptAug-PRMT                    & 96.3\% & 80.5\% & 89.4\% & 0.2684 \\
29  & Disamb2.5-PRMT&PromptAug-PRMT           & 98.8\% & 98.8\% & 98.8\% & 0.6991 \\
30  & PromptAug-PRMT&PromptAug-PRMT           & 100\% & 100\%   & 100\%  & 0.8776 \\
31  & ImageNet-IM&Disamb2.5-PRMT              & 95.4\% &  n/a      &    n/a      & 0.2919 \\
32  & Disamb-IM&Disamb2.5-PRMT                    & 98.0\% &  n/a        &  n/a        & 0.2824 \\
33  & Disamb2.5-IM&Disamb2.5-PRMT                    & 98.5\% &   n/a       &    n/a      & 0.2770 \\
34  & PromptAug-IM&Disamb2.5-PRMT                    & 96.6\% & n/a        &    n/a      & 0.2720 \\
35  & Disamb2.5-PRMT&Disamb2.5-PRMT           & 100\%  &   n/a       &     n/a     & 1.0000 \\
36  & PromptAug-PRMT&Disamb2.5-PRMT           & 99.6\% &   n/a       &    n/a      & 0.7960 \\
\hline
\multicolumn{3}{|c|}{Averages (only rows with \knn)}     & 77.1\% & 74.6\% & 77.3\% & \\
\hline

  \end{tabular}
  \caption{For all pairs of reference and query sets, we give the Centroid Accuracy, \knn Accuracy using $\knn=\{1,5\}$ and Average Cosine Similarity.   For \knn, Disamb2.5-PRMT queries are not calculated because of small sample sizes.}
  \label{table:centroid-accuracy}  
  \end{center}
\end{minipage}
\end{table}

For this study, we experimented with a large set of reference and query sets; many more than could be presented in the main text. Results are presented here.   We compute Centroid Accuracy on both images and prompts, distinguished with a -IM and -PRMT suffix respectively.  Experiments include  the image-sets: ImageNet-IM, Disamb-IM, Disamb2.5-IM, PromptAug-IM, and the prompt-sets: Disamb2.5-PRMT, PromptAug-PRMT, for a total of 36 experiments.

The calculation of Centroid Accuracy proceeds from Section~\ref{sec:low-diversity}.  For all of the images, we compute their CLIP embeddings, $E_{S,c} = ClipEmbedImage(I_{S,c})$ where $I_{S,c}$ are the images of set $S$ in class $c$.  Because CLIP is trained using cosine similarity and relies on unit vectors, to calculate the centroid location of $E_{S,c}$, $M_{S,c} = uvec(\sum_i uvec(e_{S,c}^i \in E_{S,c}))$ where $uvec(\vec{v}) = \frac{\vec{v}}{|\vec{v}|}$ is the unit vector function.    The set of class  centroids for the reference set, one per class, is $R^c$.

The nearest reference embedding in the set $R^c$ to a query embedding $q$ is $Near(q, R^c)=n \text{~where~}  n\in R^c$; randomly selected in case of ties. 

The classification correctness of an individual query, $q$, is $Z(q,R)$.  $Z(q,R) = 1.0$ when $Near(q, R^c)$ is labeled with the same class as $q$, otherwise  $Z(q,R) = 0.0$.

The Centroid Accuracy, $\CAMATH$, for a query set $Q$ and reference set $R$ is then  $\CAMATH(Q, R) = \frac{\sum_{q \in Q} Z(q,R)}{ |Q|}$.

In Section~\ref{sec:knn}, we discussed using a more general \knn approach to the zero-shot centroid based versions described earlier.    Results with $k=1,5$ are also presented in the table.  
As with Centroid Accuracy, references always use the set's training split, and queries always uses an independent evaluation split.  \knn specifics are  described in Section~\ref{sec:knn}.

Finally, in the last column, we show the average Cosine Similarity; this is the average of the cosine similarities of the query-reference embedding pairs.  These are discussed in detail in the main text, Section~\ref{sec:modified-prompts}.

\clearpage
\section{Appendix: Explicit Control of Diversity and Mean-Shift through Prompt-Tuning}
\label{appendix:nuance}

\begin{figure}[h]
\small
    \centering
    \begin{tabularx}{\textwidth}{r|c|r}
    \textbf{Query} & \textbf{Sample Images} & \textbf{Accuracy} \\

    \rotatebox[origin=c]{90}{\begin{minipage}{0.08\textwidth}\centering
    school\\bus
   \end{minipage}} &    
    \raisebox{-0.45\height}{\includegraphics[height=1.5cm]{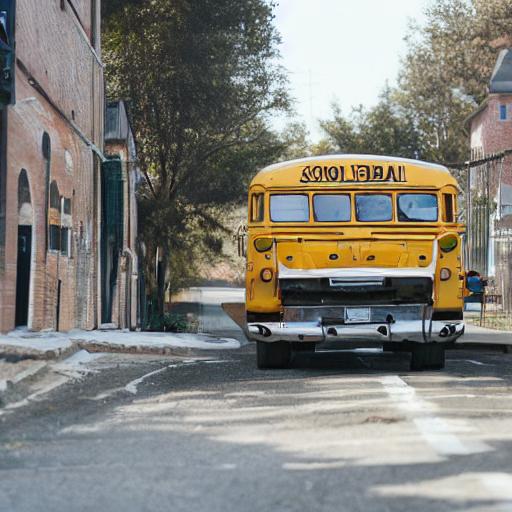}}
    \raisebox{-0.45\height}{\includegraphics[height=1.5cm]{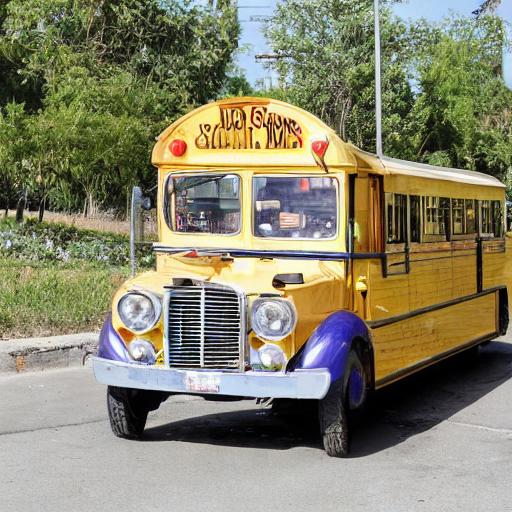}}
    \raisebox{-0.45\height}{\includegraphics[height=1.5cm]{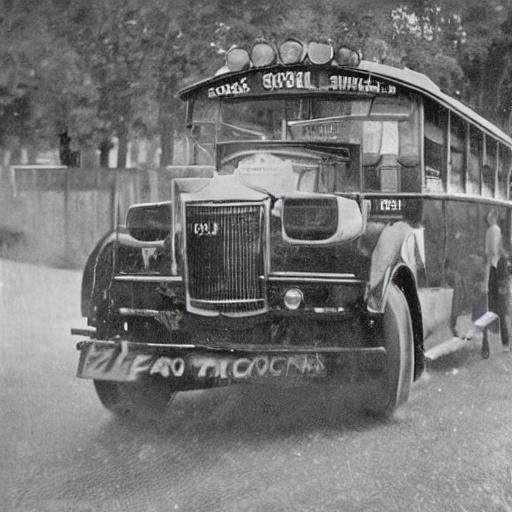}} &
    92.5\% \\

    \rotatebox[origin=c]{90}{\begin{minipage}{0.08\textwidth}\centering
    yellow\\school\\bus
    \end{minipage}} &
    \raisebox{-0.45\height}{\includegraphics[height=1.5cm]{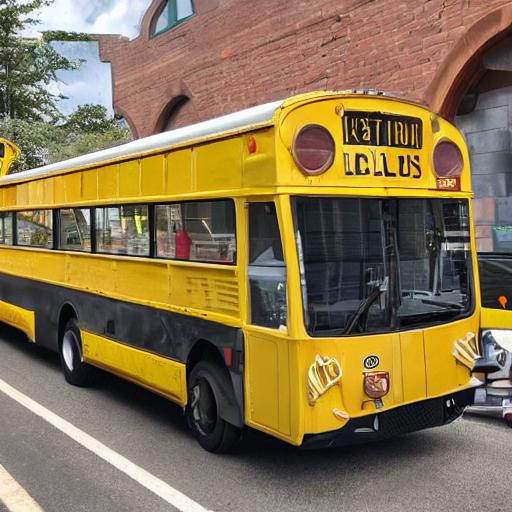}}
    \raisebox{-0.45\height}{\includegraphics[height=1.5cm]{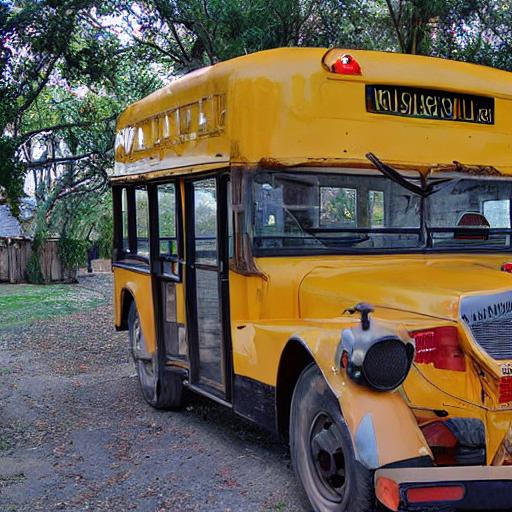}}
    \raisebox{-0.45\height}{\includegraphics[height=1.5cm]{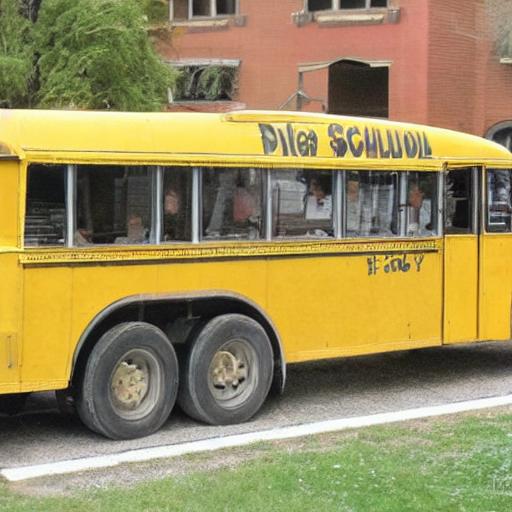}} &
    96.2\% \\

    \rotatebox[origin=c]{90}{\begin{minipage}{0.08\textwidth}\centering
    purple\\school\\bus
    \end{minipage}} &
    \raisebox{-0.45\height}{\includegraphics[height=1.5cm]{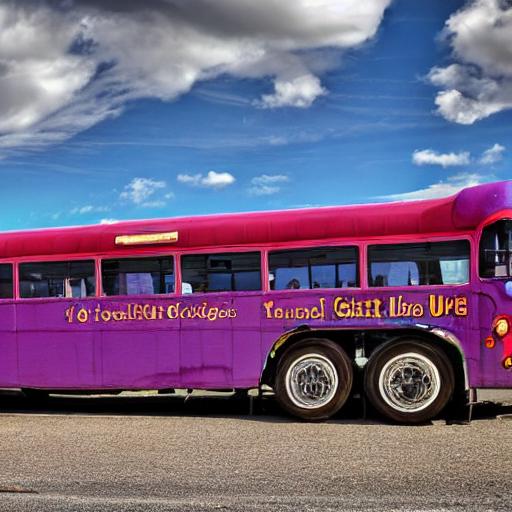}}
    \raisebox{-0.45\height}{\includegraphics[height=1.5cm]{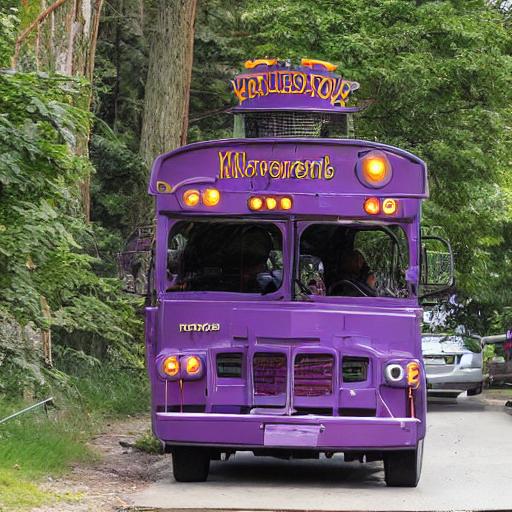}}
    \raisebox{-0.45\height}{\includegraphics[height=1.5cm]{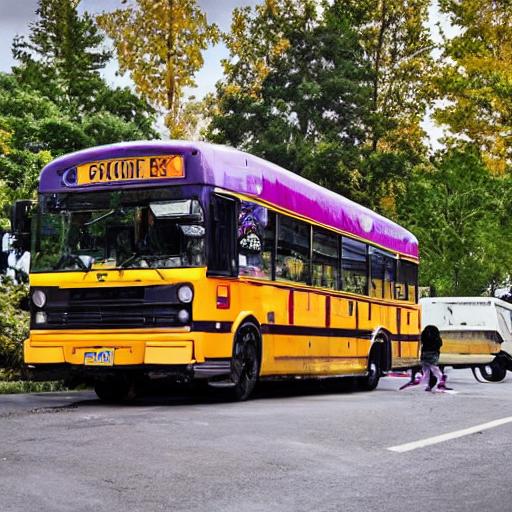}} &
    90.7\% \\

    \rotatebox[origin=c]{90}{\begin{minipage}{0.08\textwidth}\centering
    purple school\\bus in a\\jungle
    \end{minipage}} &
    \raisebox{-0.45\height}{\includegraphics[height=1.5cm]{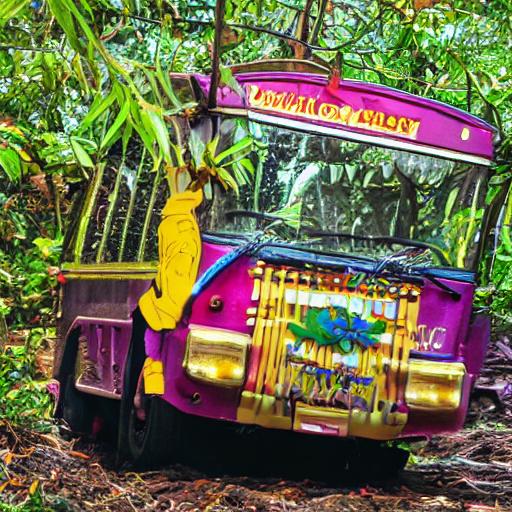}}
    \raisebox{-0.45\height}{\includegraphics[height=1.5cm]{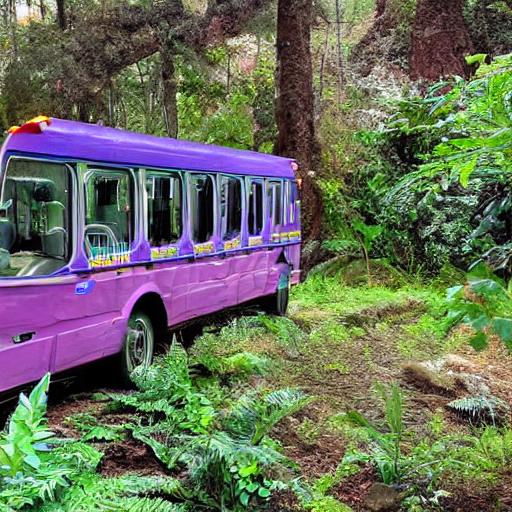}}
    \raisebox{-0.45\height}{\includegraphics[height=1.5cm]{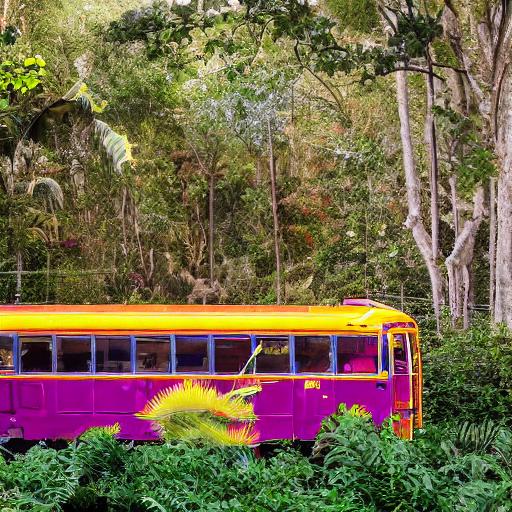}} &
    88.2\% \\
    \end{tabularx}
\\
    \caption{Centroid Accuracy for the query using ``school bus'' as the reference.}
    \label{fig:school-bus}
\end{figure}

We briefly examine the common practice (from both academia and end-users of Stable Diffusion) of fine-tuning prompts to produce broader or narrower sets of images.  We examine the effects that common prompt tuning has on variance and \meanshift\space using the prompt ``school bus'' and three modified school bus queries.  
For each of the four prompts (shown in Figure~\ref{fig:school-bus}), we synthesize 1200 new images; these will be used as the query images.  The reference set, which we hold constant, is Disamb2.5-IM; this makes the experiments comparable to Experiment 15 (shown in Appendix~\ref{appendix:centroid-accuracy}.  As a reminder, recall that ``school bus'' is a class in ImageNet (and therefore in Disamb2.5).

The new set of images created with the base prompt ``school bus'' is first tested.  Like Experiment 15, this measures only the diversity, entirely due to diffusion since the query text has not changed, and results in Centroid Accuracy of 92.5\%.   

Next, we attempt to explicitly reduce diversity by constraining the color.  The query ``yellow school bus'' reduces the diversity of school buses generated.  Since yellow is the color most often associated with school buses, this will likely have minimal impact on the centroid of the synthetic images, but rather serve primarily to reduce diversity.   The increased accuracy of 96.2\% corroborates this. Measuring the \cd produces $10.96{\times}10^{-4}$ for ``school bus'' and $5.33{\times}10^{-4}$ for ``yellow school bus'', verifying the reduced diversity of the latter.

Next, we shift the mean of the synthetic images by generating uncommon school buses, \eg ``purple school buses.'' 
Earlier,  we suggested the diffusion process is heavily influenced by the underlying image distributions of the training data; the more common the object, the more likely it is to be produced.  Since purple school buses are \emph{less likely} to be encountered than yellow, they appear less frequently in the reference set, and  therefore may be less accurately classified when used as queries. With respect to the reference school bus images, we   expect a shift in the centroid of the synthetic images (\meanshift). Correspondingly, we see reduced classification accuracy (90.7\%).   Finally, we can further increase \meanshift\space by additionally constraining the background to an unlikely scenario:  ``purple school bus in a jungle,'' this further reduces accuracy (88.2\%).  Additionally, this prompt changes the diversity by limiting the backgrounds generated.

\begin{figure}[h]
\centering
\small
\begin{center}
    \begin{tabular}{cc}
    \textbf{Set} & \textbf{Sample Images}\\

    Disamb2.5 &
    \raisebox{-0.45\height}{\includegraphics[width=0.08\textwidth]{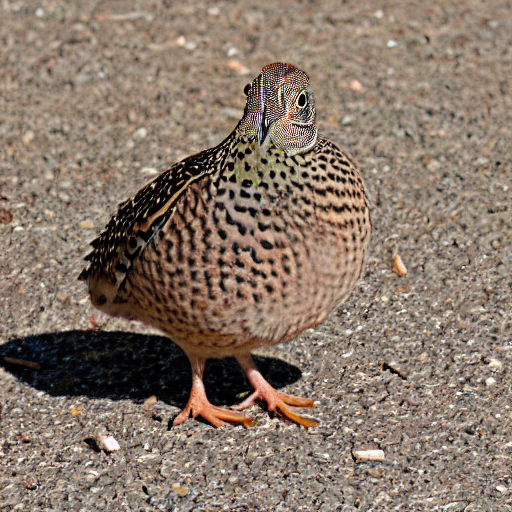}}
    \raisebox{-0.45\height}{\includegraphics[width=0.08\textwidth]{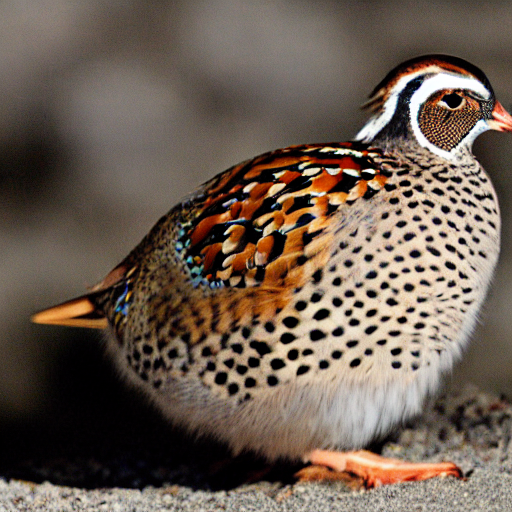}}
    \raisebox{-0.45\height}{\includegraphics[width=0.08\textwidth]{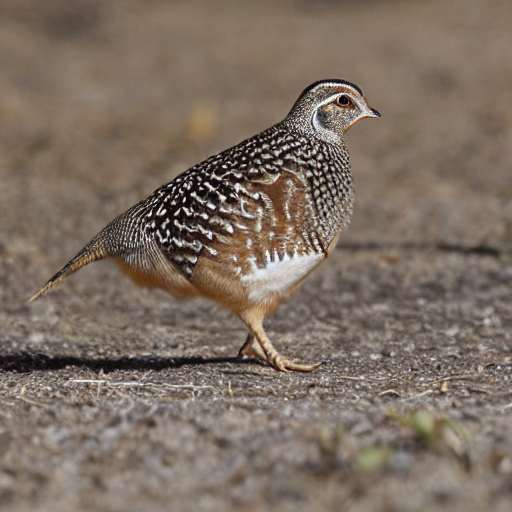}}\\

    PromptAug & 
    \raisebox{-0.45\height}{\includegraphics[width=0.08\textwidth]{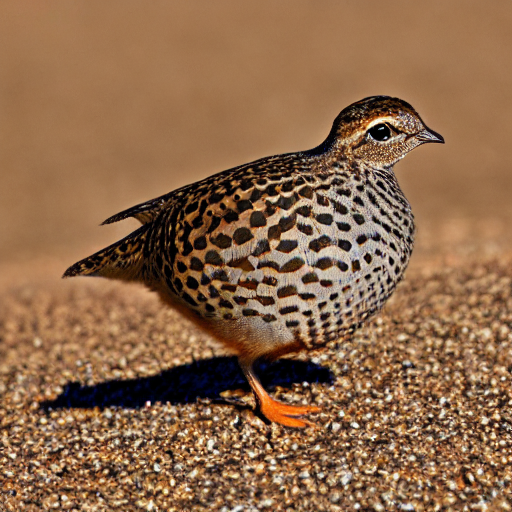}}
    \raisebox{-0.45\height}{\includegraphics[width=0.08\textwidth]{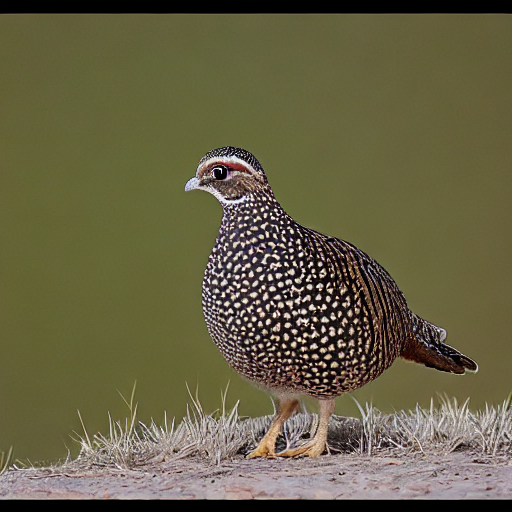}}
    \raisebox{-0.45\height}{\includegraphics[width=0.08\textwidth]{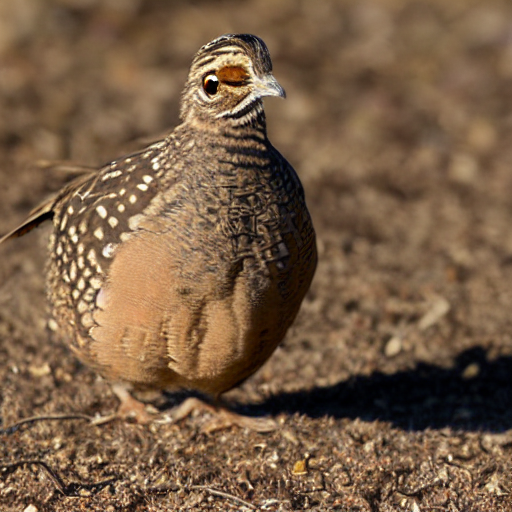}}\\
    \end{tabular}

~\\
~\\

    1. beautiful, common, extremely small quail, with many other other objects visible. Hyper-sharp.\\
    2. old, common, extremely large size quail, centered in the image. Typical snapshot.\\    
    3. ugly, extremely uncommon, slightly small size quail, centered in the image. Hyper-sharp.\\
    \caption{ Top row all generated with prompt "quail".  Bottom row, generated with the prompts shown.  There is some added diversity compared to the top row, but parts of the prompts are not captured in the image generation.}
    \label{fig:quail}
    
\end{center}
\end{figure}

We would be remiss in presenting these results without discussing pragmatic limitations of prompt tuning.  Recall that in our set with augmented prompts, PromptAug, we generated a unique prompt for each image generated.  The modified prompts used a variety of prefix and postfix modifiers to the class label.   In general, these did improve the diversity of the images generated. However, for some classes, not all of the modifiers in the prompts were present. Instead, only similar images were created, despite the modifiers that were used.  See Figure~\ref{fig:quail}.  This is commonly witnessed by end-users who carefully tune prompts to find the words and phrases that ``make it to'' the image.

\clearpage

\bibliography{refs}

\end{document}